\documentclass{article}
\usepackage{ijcai23}

\usepackage{amssymb}
\usepackage{colortbl}  %彩色表格需要加载的宏包
\usepackage{xcolor}
\usepackage{array}
\usepackage{subcaption}
\usepackage{multirow}

\usepackage[font={small}]{caption}  % 全局修改caption字号，此时为五号字体
\usepackage{pifont}       % \ding{xx}
\usepackage{bbding}       % \Checkmark,\XSolid,... (需要和pifont宏包共同使用)
\usepackage{fontawesome}  % \faCheck,\faTimes
\usepackage[section]{placeins}
\usepackage{xcolor}

\usepackage{times}
\usepackage{soul}
\usepackage{url}
\usepackage[hidelinks]{hyperref}
\usepackage[utf8]{inputenc}
\usepackage{graphicx}
\usepackage{amsmath}
\usepackage{amsthm}
\usepackage{booktabs}
\usepackage{algorithm}
\usepackage{algorithmic}
\usepackage[switch]{lineno}
\usepackage{cleveref} 
\usepackage{verbatim}
\title{Multi-view Distillation based on Multi-modal Fusion for Few-shot Action Recognition(CLIP-$\mathrm{M^2}$DF) }

\author{
Fei Guo$^1$ \and
YiKang Wang$^1$\and
Han Qi$^1$\and
WenPing Jin$^1$ \And
Li Zhu$^{*1}$
\affiliations
$^1$School of Software, Xi'an Jiaotong University\\
\emails
\{co.fly, funnyq\}@stu.xjtu.edu.cn,
\{qihan19, jinwenping\}@stu.xjtu.edu.cn,
zhuli@xjtu.edu.cn
}
\begin{document}

\maketitle
%%%%%%%%%%%%%%%%%%%%%%%%%%%%%%%%%%%%%%%%%%%%%%%%%%%%%%%%%%%%%%%%%%%%%%%%%%%%%%%%%%%%%%
%%%%%%%%%%%%%%%%%%%%%%%%%%%%%%%%%%%%%abstract%%%%%%%%%%%%%%%%%%%%%%%%%%%%%%%%%%%%%%%%%
%%%%%%%%%%%%%%%%%%%%%%%%%%%%%%%%%%%%%%%%%%%%%%%%%%%%%%%%%%%%%%%%%%%%%%%%%%%%%%%%%%%%%%
\begin{abstract}
In recent years, few-shot action recognition has attracted increasing attention. It generally adopts the paradigm of meta-learning. 
In this field, overcoming the overlapping distribution of classes and outliers is still a challenging problem based on limited samples. 
We believe the combination of Multi-modal and Multi-view can improve this issue depending on information complementarity. 
Therefore, we propose a method of Multi-view Distillation based on Multi-modal Fusion. 
Firstly, a Probability Prompt Selector for the query is constructed to generate probability prompt embedding based on the comparison score between the prompt embeddings of the support and the visual embedding of the query. 
Secondly, we establish a Multi-view. In each view, we fuse the prompt embedding as consistent information with visual and the global or local temporal context to overcome the overlapping distribution of classes and outliers. 
Thirdly, we perform the distance fusion for the Multi-view and the mutual distillation of matching ability from one to another, enabling the model to be more robust to the distribution bias.
Our code is available at the URL: \url{https://github.com/cofly2014/MDMF}.
\end{abstract}

%%%%%%%%%%%%%%%%%%%%%%%%%%%%%%%%%%%%%%%%%%%%%%%%%%%%%%%%%%%%%%%%%%%%%%%%%%%%%%%%%%%%%%
%%%%%%%%%%%%%%%%%%%%%%%%%%%%%%%%%%%Introduction%%%%%%%%%%%%%%%%%%%%%%%%%%%%%%%%%%%%%%%
%%%%%%%%%%%%%%%%%%%%%%%%%%%%%%%%%%%%%%%%%%%%%%%%%%%%%%%%%%%%%%%%%%%%%%%%%%%%%%%%%%%%%%
\section{Introduction}
%%%%%%%%%%%%%%%%%%%%%%%%%%%%%%%%%%%提出问题%%%%%%%%%%%%%%%%%%%%%%%%%%%%%%%%%%%%%%%%%%%%%%%%
Few-shot action recognition mainly solves two problems: 
(1) How to represent spatiotemporal sequences with distinguishability under limited samples. 
(2) How to establish a sequence comparison between support and query. 
The overlapping distribution of classes and the outliers in classes often affect the recognition accuracy, even if there is a preferred solution for the two problems.
We need to construct common representations to overcome the influence of data distribution.

%%%%%%%%%%%%%%%%%%%%%%%%%%%%%%%目前相关的解决方法%%%%%%%%%%%%%%%%%%%%%%%%%%%%%%%%%%%%%%%%%%%%
The works related to probability distribution are as follows: 
PAL\cite{zhu2021few-PAL} uses Prototype-centered attentive learning to reduce the negative impact of outlier samples.
MPRE\cite{liu2022multidimensionalMPRE} uses Prototype Aggregation Adaptive Loss, Cross-Enhanced Prototype, and Dynamic Temporal Transformation to solve the problem of distribution bias. 
FTAN\cite{yu4104257ftan} uses the joint integration of TCM and FCSM in ATA to generate better feature embedding, solving the problems of time distribution, intra-class time offset, and inter-class local similarity. 
Because the samples are limited in few-shot learning, the shuffle or the attention mechanism perhaps can not supply a superior solution. 
HyRSM \cite{wang2022-hybrid} uses the hybrid relation module and the bidirectional Mean Hausdorff Metric to overcome the data noise and distribution bias.
AMFAR\cite{wanyan2023active} estimates a specific modality's reliability based on its posterior distribution's determinacy. But, this work uses optical flow data, which is difficult to obtain.
CLIP-FSAR\cite{wang2023clip} uses the label as a supplement for visuals that can solve the problem of limited samples and uses a Transformer to get a better representation. However, it only uses the labels for support, and perhaps it has not utilized labels efficiently.

%%%%%%%%%%%%%%%%%%%%%%%%%%%%%%%%%%%%%解决的问题%%%%%%%%%%%%%%%%%%%%%%%%%%%%%%%%%%%%%%%%%%%%%
We aim to deal with the following problems:
(1) Trying to use the labels efficiently not only for support samples but also for query samples.
(2) Overcoming the inter-class distribution overlapping and outliers by fusing the prompt embedding and using the Multi-modal information efficiently.
(3) Constructing a robust few-shot model through reducing data distribution bias by the fusion and distillation of Multi-view related to different temporal contexts.
Our idea is to make the model pay attention to the universal features of data from two points: 
Multi-modal semantic supplementation and 
Multi-view temporal context of global and local features. 

%%%%%%%%%%%%%%%%%%%%%%%%%%%%%%%%%%%%方法概述%%%%%%%%%%%%%%%%%%%%%%%%%%%%%%%%%%%%%%%%%%%%%%%%
From a technical perspective, we propose the Multi-view Distillation based on Multi-modal Fusion using the CLIP\cite{Clip-origin2021} as the backbone.
%第一个点是通过文本来增加类间分布的稳定性和可区分性
(1) The label prompt describing a specific category of videos has consistency for all videos in such category. So, it is robust for the representation of videos and can contribute to the stability and distinguishability of inter-class distribution. %类间分布的 稳定性和可区分性
For CLIP-FSAR \cite{wang2023clip}, in each episode, label prompt embedding and visual embedding are concatenated for support to counteract sample-specific distributions. However, for queries, only visual embedding is used. This results in an inconsistency in the amount of information between query and support. 
We propose a Probability Prompt Selector to solve this problem. 
In an N-way K-shot setting, the category of the query must belong to the categories of the supports. We compare the visual embedding of the query with the prompt embedding of supports to obtain a set of matching scores and convert them into a probability distribution. Depending on the probability, we select the prompt embedding for the query by uniform sampling.
%第二个点不同分支的多模态融合
(2)No matter what kind of feature we use, we can not avoid the inter-class distribution bias and the outliers. 
We use the Multi-modal features from two different views to improve the issues.
A Local Temporal Context Extractor is used to encourage the propagation of local sequence information, 
also 
a Global Temporal Context Extractor is used to promote the propagation of global sequence information.
For each view, we use the Cross-Transformer to fuse the prompt embedding, local (global) context, and the standard visual features to get more distinguishable features.
%第三个点使用两个分支的蒸馏来减少离群点影响
(3) Then, we perform knowledge mutual distillation between the two views to force global and local context representation to have a consistent class prediction. According to the posterior distribution of the text-mode comparison and the visual-mode comparison, the distillation direction for each query is different.
Following these ways, the Multi-modal features could overcome the inter-class overlapping and outliers, enabling learning of universal features less affected by the special samples, and Multi-view can enhance the information complementary to the model's robustness.

%%%%%%%%%%%%%%%%%%%%%%%%%%%%%%%%%%%%%%%%%%%%%%%%%贡献%%%%%%%%%%%%%%%%%%%%%%%%%%%%%%%%%%%%%
Our contributions are summarized as follows:
(1) We first proposed a framework of the Multi-view Distillation based on Multi-modal Fusion.
(2) In the Multi-modal prototype matching paradigm based on CLIP, we propose a new concept of probability prompt embedding to compensate for the information inconsistency between prototypes and queries to utilize labels efficiently.
(3) We propose Multi-view context extractors to get the features from two views. In each view, a Cross-Transformer is used to fuse the prompt embedding and visual feature.
Then, we use distance fusion and mutual distillation between Multi-view to enhance the performance further.
(4) A large number of experimental results on five benchmarks, HMDB51, UCF101, Kinetics, and Something-to-Something-V2-Full(Small), demonstrate the rationality of our setting and the effectiveness of our proposed method. The results can be compared with the state-of-the-art methods.

%%%%%%%%%%%%%%%%%%%%%%%%%%%%%%%%%%%%%%%%%%%%%%%%%%%%%%%%%%%%%%%%%%%%%%%%%%%%%%%%%%%%%%
%%%%%%%%%%%%%%%%%%%%%%%%%%%%%%%%%Related Work%%%%%%%%%%%%%%%%%%%%%%%%%%%%%%%%%%%%%%%%%
%%%%%%%%%%%%%%%%%%%%%%%%%%%%%%%%%%%%%%%%%%%%%%%%%%%%%%%%%%%%%%%%%%%%%%%%%%%%%%%%%%%%%%
\section{Related Work}
%%%%%%%%%%%%%%%%%%%%%%%%%%%%%%%%%%%%%%%%%%%%%%%%%%%%%%%%%%%%%%%%%%%%
\subsection{Few-shot Image Classification}
Few-shot image classification aims to identify objects of unseen categories using only a few labels and also will use a large number of samples under the seen categories.
Research in this field is broadly divided into three categories: metric-based, optimization-based, and augmentation-based. 
Metric-based methods 
\cite{simon2020adaptive-few_shot_learning1,snell2017prototypical_few_shot_learning2,sung2018learning-relation-networkfew_shot_learning3} extract the spatiotemporal features and use support-query matching rules to classify the query. The matching rules contain $cosine$ similarity, Euclidean distance, and learnable distance based on neural networks. 
Optimization-based methods make the provided model well-initialized and easy to reach the optimal point, just as MAML\cite{finn2017model-MAML-few-shot-model1} and related variants. 
Augmentation-based methods 
\cite{chen1804semantic-few-augmentation3,ratner2017learning-few-augmentation2,perez2017effectiveness-few-augmentation1} make use of generative strategies that could produce lots of valuable data under conditions without enough data. 
%%%%%%%%%%%%%%%%%%%%%%%%%%%%%%%%%%%%%%%%%%%%%%%%%%%%%%%%%%%%%%%%%%%%%%%
\subsection{Few-shot Action Recognition}
Unlike image classification, action recognition must consider the temporal dimension and the extraction of keyframes. Few-shot action recognition aims to solve the unrealistic problem of getting lots of labeled video samples. There are several important works in this field.
CMN\cite{zhu2018compound} adopts a memory network to store the representations and classify the action videos by matching and sorting. 
OTAM\cite{cao2020few-otam} calculates the distance matrix of frames based on the DTW \cite{muller2007dynamic} method and performs strict matching. 
TRX\cite{perrett2021temporal-trx} uses a subsequence cross-attention to extract feature prototypes on different temporal scales, which can effectively alleviate the temporal misalignment.
STRM \cite{thatipelli2022spatio-strm} adds some pre-processing for the feature enrichment for TRX.
MTFAN\cite{wu2022motion} proposes an end-to-end network by jointly exploring task-specific motion modulation and multi-level temporal fragment alignment.
MoLo\cite{wang2023molo} develops a motion-augmented long-short contrastive learning method to jointly model the global contextual information and motion dynamics. 
CLIP-FSAR\cite{wang2023clip} leverages the strong generalization ability of CLIP \cite{Clip-origin2021} trained on hundreds of millions of datasets and uses encoders to encode text and images, then gets the enhancement of Transformer for comparison.
AMFAR \cite{wanyan2023active} uses bidirectional distillation to capture differentiated task-specific knowledge from reliable modalities to improve the representation of unreliable modalities.
Similar to CLIP-FSAR, our work is also based on CLIP.
%%%%%%%%%%%%%%%%%%%%%%%%%%%%%%%%%%%%%%%%%%%%%%%%%%%%%%%%%%%%%%%%%%%
\subsection{Knowledge Distillation}
Knowledge distillation is the classic model compression method, with the core idea of guiding lightweight student models to mimic better-performing and more structurally complex teacher models. 
Optimization strategies, such as mutual learning and self-learning through neural networks and
data resources, such as unlabelled and cross-modal, significantly enhance model performance. 
Knowledge Amalgamation\cite{2019Amalgamating} is the migration of multiple tasks into a single student model to make it capable of handling multiple tasks. 
Mutual distillation\cite{2018Deep} addresses using student models to learn from each other to improve performance without a robust teacher network, avoiding the reliance on large-scale teacher models.
\cite{2018Emotion} proposed cross-modal affective recognition with the distillation of data features from different modalities. 
AMFAR \cite{wanyan2023active} is related to the mutual distillation of Multi-modal. Our work is based on the Multi-view distillation.

%%%%%%%%%%%%%%%%%%%%%%%%%%%%%%%%%%%%%%%%%%%%%%%%%%%%%%%%%%%%%%%%%%%%%%%%%%%%%%%%%%%%%%
%%%%%%%%%%%%%%%%%%%%%%%%%%%%%%%%%%%%%%Method%%%%%%%%%%%%%%%%%%%%%%%%%%%%%%%%%%%%%%%%%%
%%%%%%%%%%%%%%%%%%%%%%%%%%%%%%%%%%%%%%%%%%%%%%%%%%%%%%%%%%%%%%%%%%%%%%%%%%%%%%%%%%%%%%
\section{Method}

\subsection{Problem Setting}
In the field of few-shot action recognition, the video datasets are split into $D_{train}$, $D_ {test}$, $D_ {val}$, all the split datasets should be disjoint, which means there are no overlapping classes between each split dataset. In the $D_{train}$, it contains abundant labeled data for each action class, while there are only a few labeled samples in the $D_ {test}$, and the $D_ {val}$ is used for model evaluation during the training episode. No matter $D_{train}$, $D_ {test}$, or $D_ {val}$, they all follow a standard episode rule. The episode, also called a task, occurs during the training, testing, or validation. In each episode, $N$ classes with $K$ samples in $D_{train}$, $D_ {test}$, or $D_ {val}$ are sampled as ``support set". The samples from the rest videos of each split DB are sampled as ``query set", just as $P$ samples are selected from $N$ classes to construct the ``query set". 
The goal of few-shot action recognition is to train a model using $D_{train}$, which can be generalized well to the novel classes in the $D_ {test}$ only using $N\times K$ samples in the support set $D_ {test}$. Let $q = ({q_1, q_2, \cdots , q_m })$ represents a query video with $m$ uniformly sampled frames. We use $C =\{c_1,\cdots, c_N\}$ to represent the class set, and we aim to classify a query video $q$ into one of the classes $c_i \in C$. In our work, the support set is defined as $S$, and the query set is defined as $Q$. 
For the class $c$, the support set $s_c$  can be expressed as $s_c = \{s_c^1, \cdots ,s_c^k \cdots, s_c^K\}, 1 \leq k \leq K$, and $s_c^k= ({s_c^{k,1}, \cdots, s_c^{k,m}}),  1 \leq k \leq K$, $m$ is the frame number.

\begin{figure*}
\centering
\includegraphics[width=15.3cm]{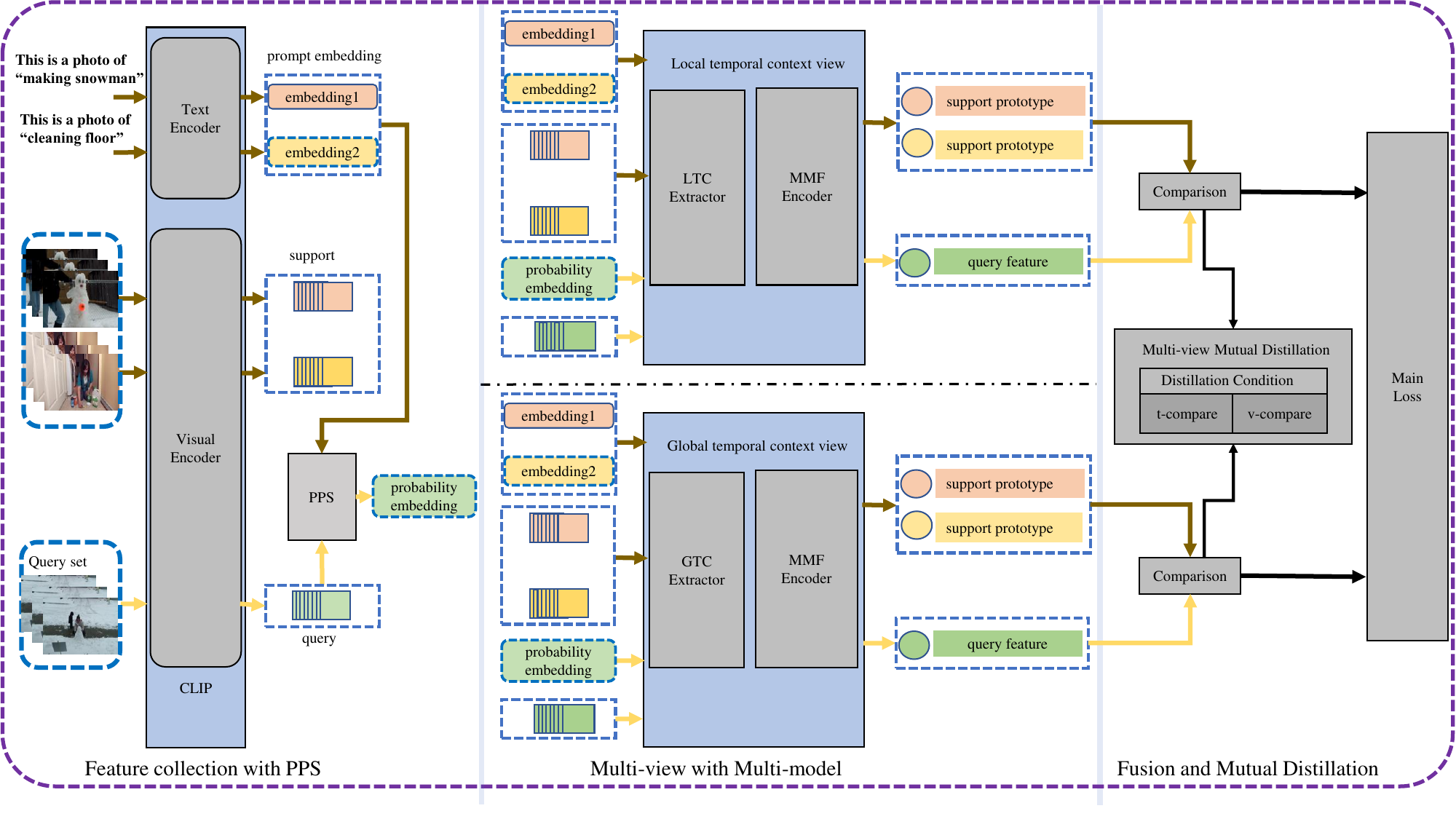}
\caption{\label{fig:1} The framework. The grey regions are the modules or some simple operations. The blue dashed regions represent the features. 
The khaki arrows are the data flow of support, including the visual and label. 
The yellow arrows are the data flow of the query, including the visual and label. 
There are 6 modules in our model. 
(1) CLIP(Visual Encoder and Text Encoder).
(2) Probability Prompt Selector (PPS). 
(3) Local Temporal Context Extractor (LTCE). 
(4) Global Temporal Context Extractor (GTCE). 
(5) Multi-modal Fusion Encoder (MMFE). 
(6) Multiple-view Mutual Distillation(MVMD). 
There are some other operations:
(a) Distillation Condition.
(b) Comparison.
(c) Main loss
} 
\vspace{-0.8em}
\end{figure*}
%%%%%%%%%%%%%%%%%%%%%%%%%%%%%%%%%%%%%%%%%%%%%%%%%%%%%%%%%%%%%%%%%%%%%%%%%%%%%%%%%%%%%%
\subsection{Overview}
The framework of our model can be seen in \cref{fig:1}.
\textbf{Firstly}, the visual encoder of CLIP is used to get the visual embedding of support and query. The text encoder of CLIP  is used to get the label prompt embedding of the support categories in each episode. Also, a Probability Prompt Selector (PPS) is proposed to generate a probability prompt for each query video. The prompt embeddings introduce the stable feature that does not change with sample distribution and ensure the consistency of information between the representation of support and query.
\textbf{Secondly}, we introduce the Multi-view structure, where each view fuses the Multi-modal related to the label feature and visual feature. The Multi-view structure is as follows: 
(1) Local Temporal Context Extractor (LTCE). Using several Conv1d operations in the temporal dimension, each frame could contain the context information of adjacent frames.
(2) Global Temporal Context Extractor (GTCE). Using the TCN \cite{lea2016temporal-tcn} network in the temporal dimension, each frame could get the global sequence context.
(3) Multi-modal Fusion Encoder (MMFE). The core of MMFE is a Cross Transformer, which is introduced for each view to extract Multi-modal features related to the label prompt and visual. 
This module concatenates each video context from the LTCE (or GTCE) with prompt embedding as the $Query$ 
and concatenates the features from the CLIP visual encoder with prompt embedding as the $Key$ and $Value$. The fused features are then obtained through the Cross Transformer.
\textbf{Thirdly}, fusion and distillation of the two views enable the model to register the Multi-modal features from global and local temporal contexts, thereby enabling the model to learn more general features from the data. 
%%%%%%%%%%%%%%%%%%%%%%%%%%%%%%%%%%%%%%%%%%%%%%%%%%%%%%%%%%%%%%%%%%%%%%%%%%%%%%%%%%%%%%
\subsection{Probability Prompt Selector(PPS)}
The existing CLIP-based few-shot action recognition works, just as CLIP-FSAR \cite{wang2023clip} and MORN \cite{ni2022Multi-modal-CLIP} incorporate text information while constructing support prototype. However, the query still maintains the mono-modality of visual. By intuition, the amount of information between the support prototype and the query is inconsistent. 
Because the label prompt embedding under the same category has feature consistency for all videos, it will not change for different video instances and is robust to the probability distribution of representation. So, it is essential to introduce label prompt embedding into the query. 
In the paradigm of meta-learning, although there is no query label beforehand, it must belong to the label set of the support in each episode. 
Given a query $q$, we assume the representation from the visual encoder is $f_{q}$, and label prompt embedding of support class $c$ from the text encoder is $token_c$, then we calculate the $cosine$ similarity between $f_{q}$ and $token_c$:
\begin{equation}  \label{eq1}
   sim^{q}_{c}= \frac{<f_{q}, token_{c}> }{|f_{q}| \cdot |token_{c}|}
\end{equation}
Then, we use the Softmax with the temperature coefficient $t$ to transfer the similarity value into a probability distribution.
\begin{equation}  \label{eq2}
    prob^{q}_{c}= \frac{exp(sim^{q}_{c}/t) }{ \sum\limits_{c{'} \in C} exp(sim^{q}_{c'}/t)}
\end{equation}
where $C =\{c_1,\cdots, c_N\}$ is the category set in each episode.
According to the probability distribution, we use uniform sampling to sample the prompt embedding in the label set of support for the query $q$. Through the PPS, we have the probability embedding for the query. 

However, the video-text matching is still not accurate enough for video. We ultimately need to rely on the comparison between query and prototype. 
Extracting more information from frame sequences and integrating the label prompt information into visual information is essential. 

%%%%%%%%%%%%%%%%%%%%%%%%%%%%%%%%%%%%%%%%%%%%%%%%%%%%%%%%%%%%%%%%%%%%%%%%%%%%%%%%%%%%%%
\subsection{Multi-view Structure} \label{section: Multi-view}
Our work considers the fusion of Multi-modal features from two views. 
The first view is the local temporal context in the temporal dimension, and
the second view is the global temporal context in the temporal dimension. 
%%%%%
\subsubsection{Local Temporal Context Extractor(LTCE)}
The \cref{fig:2} (Left) depicts the LTCE. 
This Extractor includes a series of operations such as Conv1d, Relu, BN, etc. After these operations, the features pay more attention to the information of adjacent frames in front or after, thus getting the local temporal context of the features. 
For each video, we select 8 frames as the full sequence, and we use a Convolution kernel with a size equal to 3 to get the local temporal context. 
Given the features $F=[f^1,f^2..., f^T] \in R^{T \times D}$ from the CLIP visual encoder, they are operated by the LTCE. \cref{eq3} shows the operations in the LTCE.
\begin{small}
\begin{equation}  \label{eq3}
  \begin{split}
    &F^{1} = F*W_1, F^{2} = RELU(F^{1}),F^{3} = BN(F^{2})\\
    &F^{4} =F^{3}*W_2,F^{5} = RELU(F^{4}),F^l = BN(F^{5})\\
  \end{split}
\end{equation}
\end{small}
where $W_1$ and $W_2$ are the Convolution kernels. $F^{i}, i \in \{1,...,5\}$ is the temporary variable.
%%%%%%
\subsubsection{Global Temporal Context Extractor(GTCE)}
Given the features $F=[f^1,f^2..., f^T] \in R^{T \times D}$ from the CLIP visual encoder. We use TCN\cite{lea2016temporal-tcn} to extract the global temporal features. For details, we use a TCN with three layers to get the temporal features of frames. Because the TCN uses Dilated Convolution and Causal Convolution, the dilated rate grows exponentially by 2. When $T$ equals 8, the last frame of the output feature can capture the temporal context for the full sequence. In other words, the frame number of the output is the same as the input, and the last frame pays attention to the features in a time range from 1 to $T$.
We copy the last frame of the output for $T$ copies and add the input as the global temporal context. See the \cref{eq4}.

\begin{small}
\begin{equation}  \label{eq4}
    F^g = Repeat(TCN(F)[-1]) + F
\end{equation}
\end{small}
%%%%%%

%%%%%%%%%%%%%%%%%%%%%%%%%%%%%%%%%%%%%%%%%%%%%%%%%%%%%%%%%%%%%%%%%%%%%%%%%%%%%%%%%%%%%%
\subsection{Multi-modal Fusion Encoder (MMFE)}
\begin{figure}[ht]
\centering
\includegraphics[width=7.8cm]{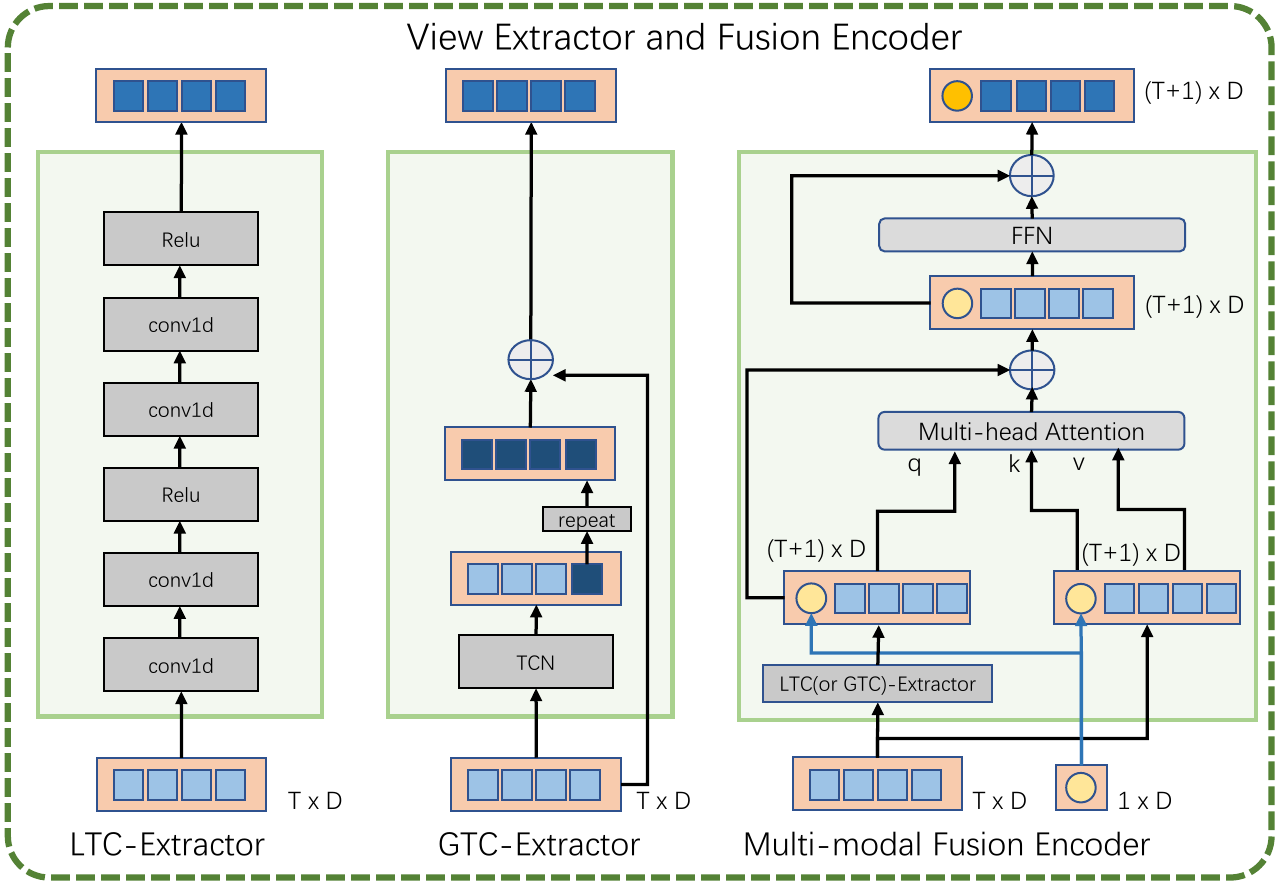}
\caption{\label{fig:2} 
The left part is the LTCE. It contains several Conv1d operations. According to Conv1d, the features focus on the local temporal context. 
The middle part is the GTCE. The core part is TCN. We add the last frame of the output to each input frame. The features focus on the global temporal context. 
The right part is the Cross Transformer. Features that are from the LTCE (GTCE) and features that are directly from the CLIP visual encoder are concatenated with the prompt embedding. The features from LTCE (GTCE) are used as the Queries, and the features from the CLIP visual encoder are used as the Keys and Values.
} 
\vspace{-1em}
\end{figure}

In \cref{section: Multi-view}, we get different temporal contexts from two views.
Now, we will study how to fuse the original sequence with the local temporal context(or global temporal context) and the corresponding prompt embedding.
The operations for the local and global temporal views are similar. \cref{fig:2} (Right) shows the Multi-modal Fusion Encoder using a Cross-Transformer.
%%%%%%%
\subsubsection{Multi-modal Feature Concatenation} 
In our work,
given the features $F = [ f^1, f^2, ..., f^T] $ from the CLIP visual encoder,
the temporal context feature $F^{b} = [f^{b,1}, f^{b,2}, ..., f^{b,T}]$ from the LTCE (or GTCE), and the corresponding $token$ (prompt embedding) from the CLIP text encoder or PPS,  
we concatenate the prompt embedding of the text encoder with the visual features of support and the local (or global) temporal context of support, respectively.
We do the same operation for the probability prompt embedding from PPS, visual features of query, and the local (or global) temporal context of query. 
as the \cref{eq5} and \cref{eq6} 
\begin{small}
\begin{equation}  \label{eq5}
  \hat{F} = Contact(token, F)
\end{equation}
\end{small}
\vspace{-1.5em}
\begin{small}
\begin{equation}  \label{eq6}
  \hat{F^{b}} = Contact(token, F^{b})
\end{equation}
\end{small}
Note: 
$b$ means the view, which can be $g$ or $l$ as the global or local temporal context. See in the \cref{section: Multi-view}.
%%%%%%%
\subsubsection{Multi-modal Feature Extraction}
We use $\hat{F^{b}}$ as the $Query$, and $\hat{F}$ as the $Key$ and $Value$, and send them into a Cross Transformer. 
\begin{small}
\begin{equation}  \label{eq7}
   \widetilde{F^b} = 
   \mathbb{T}
   (\hat{F^{b}} + f_{pos},\hat{F} + f_{pos},\hat{F} + f_{pos}) 
\end{equation}
\end{small}
where $\mathbb{T}$ is the Transformer that contains the Multi-head Attention and FFN. $f_{pos} \in R^{(T+1) \times D}$ means the position embeddings to encode the position. $\widetilde{F^b} \in R^{(T+1) \times D}$.

\subsubsection{Loss of Temporal Context View} \label{Loss of Temporal Context view}
For every view through the MMFE, in each episode, we assume the sample feature under support class c as $\widetilde{F}^b_{s_c^k} $ where $k \in \{1,..., K\}$ and the feature of a query as $\widetilde{F}^b_{q}$. Here, $b$ can be $g$ or $l$ as the global or local temporal context view.
Using the average aggregation for the support features, the prototype is calculated as follows: 
\begin{small}
\begin{equation} \label{eq8}
      U^b_c = \frac{1}{K}\sum_{k=1}^{K}\widetilde{F}^b_{s_c^k}
\end{equation}
\end{small}
We only use the visual frames (not including the first item of the query and the support prototype) from the MMFE to calculate the distance.
We calculate the distance of the global context view as the $dis(\widetilde{F^g_{q}}, U^g_c)$ and the distance of the local context view as $dis(\widetilde{F^l_{q}}, U^l_c)$.
\begin{small}
\begin{equation} \label{eq9}
    dis_{total}(q, s_c) = dis(\widetilde{F^g_{q}}, U^g_c) +  dis(\widetilde{F^l_{q}}, U^l_c)
\end{equation}
\end{small}
Using Softmax for $dis_{total}(s_c,q)$, we can get the classification probability, see \cref{eq16}. We assume $L_{main}$ is the Cross-Entropy Loss between the probability and the ground truth. 

%%%%%%%%%%%%%%%%%%%%%%%%%%%%%%%%%%%%%%%%%%%%%%%%%%%%%%%%%%%%%%%%%%%%%%%%%%%%%%%%%%%%%%
\subsection{Multiple-view Mutual Distillation (MVMD)}
%%%%
\subsubsection{Distillation conditions}
Inspired by the work \cite{wanyan2023active}, 
we select queries with significant differences in reliability between the two views, 
where the more reliable view is regarded as the primary view. 
We define that the reliable view of each query should reflect more discriminative features of specific tasks, so it deserves more attention in few-shot learning. 
For a query, the reliable view may vary across different tasks, as the contribution of a specific view largely depends on the context information of the query and support in each episode.
In the \cref{fig:1}, we can see the output of the Local (Global) Temporal Context view contains two modes: visual mode and text mode. Our distillation conditions are based on the view-specific posterior distribution for both the visual embedding comparison and the label prompt embedding comparison.
Given a query $q$ and its probability prompt label $lable_q$, the view-specific posterior distributions of two modes are as follows:
\begin{small}
\begin{equation} \label{eq10}
\mathcal{P}^b(y^{visual}_q=c| q, label_q) = \frac{exp(-dis(\widetilde{F^b_{q}}, U^b_c))}
                       {\sum\limits_{c^{'} \in C}exp(-dis(\widetilde{F^b_{q}}, U^b_{c'}))}
\end{equation}
\end{small}
\vspace{-0.4em}
\begin{small}
\begin{equation} \label{eq11}
 \mathcal{P}^b(y^{text}_q=c|  q, label_q) = \frac{exp(cosin(token^b_{q}, Token^b_c))}
                       {\sum\limits_{c^{'} \in C}exp(cosin(token^b_{q}, Token_{c'}^b))  }
\end{equation}
\end{small}
Similar to \cref{Loss of Temporal Context view}, $b$ can be $g$ or $l$ as the global or local temporal context view.  
The $visual$ means the visual mode and $text$ means the text mode.
We define 
$token^b_{q}$ as the first item of $\widetilde{F^b_{q}}$ 
and 
$Token^b_c$ as the first item of $U^b_c$ 
where $c\in [c_1,...c_N]$. $Token^b_c$ is prompt embedding for the prototype $U^b_c$ which is belongs to category $c$ and the view $b$.

We define the distillation discriminant score as the maximum element of the view-specific posterior distribution.

\begin{figure}[ht] 
\centering
\includegraphics[width=7.3cm]{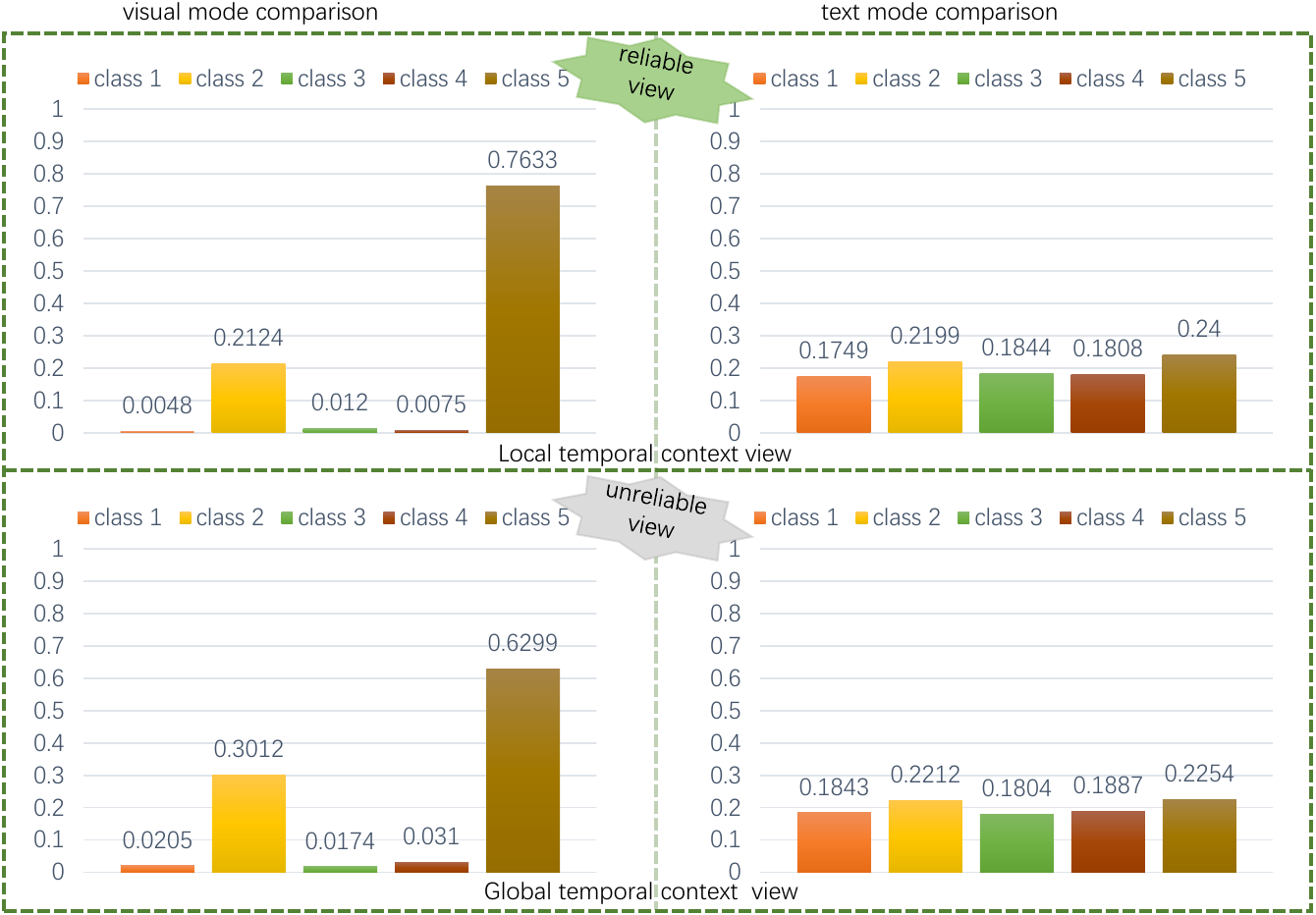}
\caption{\label{fig:3}
The distillation condition. 
The up is the local temporal context view for visual comparison and text comparison. 
The down is the global temporal context view for visual comparison and text comparison. 
We can see the $0.7633>0.6299$ and $0.24>0.2254$, so the local temporal context view is reliable.
} 
\vspace{-1.0em}
\end{figure}

\begin{small}
\begin{equation} \label{eq12}
  \hat{c}^{b}_q = \mathop{max}\limits_{k}\mathcal{P}^b(y^{visual}_q=k|  q, label_q))
\end{equation}
\end{small}

\begin{small}
\begin{equation} \label{eq13}
  \tilde{c}^{b}_q = \mathop{max}\limits_{k}\mathcal{P}^b(y^{text}_q=k|  q, label_q))
\end{equation}
\end{small}
where for the query $q$, $\hat{c}^{b}_q$ is the discriminant score of visual mode, and $\tilde{c}^{b}_q$ is discriminant score of text mode.

For the query, if a specific view achieves a higher discriminant score both for visual mode and text mode. This view is reliable for expressing discriminative action features in each episode. 
On the contrary, if the discriminant score is lower in both modes, the view may be less reliable in identifying actions. See the \cref{fig:3}, and the local temporal context view is reliable. We define the set of global view reliable samples as $\Omega^g$ and the set of local view reliable samples as $\Omega^l$.

\subsubsection{Mutual Distillation}
Depending on the $ \hat{c}^{b}_q$ and $ \tilde{c}^{b}_q$ where $b \in \{g,l \}$,  KL divergence is used for the mutual distillation. For each query, the reliable view acts as the teacher, and the unreliable view acts as the student.
Mutual distillation Losses are as follows:
\begin{small}
\begin{equation} \label{eq14}
\begin{split}
  &L_{g\rightarrow l} =  \frac{1}{\mathop{\sum} \limits_{ q \in \Omega^g}  (\hat{c}^{g}_q + \tilde{c}^{g}_q)}
                       \mathop{\sum} \limits_{ q \in \Omega^g}  (\hat{c}^{g}_q + \tilde{c}^{g}_q)D_{KL}(P^g_q, P^l_q)
                       \\
  &L_{l\rightarrow g} =  \frac{1}{\mathop{\sum} \limits_{ q \in \Omega^l}  (\hat{c}^{l}_q + \tilde{c}^{l}_q)}
                       \mathop{\sum} \limits_{ q \in \Omega^l}  (\hat{c}^{l}_q + \tilde{c}^{l}_i)D_{KL}(P^l_q, P^g_q)
                       \\
\end{split}
\end{equation}
\end{small}
where $P^l_q$ and $P^g_q$ are calculated as the \cref{eq10} for classification distribution of visual mode.

The final loss can be denoted as:
\begin{small}
\begin{equation} \label{eq15}
    L=   L_{main} + \lambda( L_{g\rightarrow l} + L_{l\rightarrow g})
\end{equation}
\end{small}
where $\lambda$ is the hyper-parameter. 

%%%%%%%%%%%%%%%%%%%%%%%%%%%%%%%%%%%%%%%%%%%%%%%%%%%%%%%%%%%%%%%%%%%%%%%%%%%%%%%%%%%%
\subsection{Inference}
In the meta-testing stage, we fuse the distance of the local temporal context view and the distance of the global temporal context view for inference. See the \cref{eq9}.
Using Softmax for the $dis_{total}$ of all the support prototypes in each episode, we can get the classification probability for inference.
\begin{small}
\begin{equation}  \label{eq16}
    \setlength\belowdisplayskip{-1pt}
    P(y=c|q)=\frac{exp(dis_{total}(s_c, q))} {\sum\limits_{c^{'} \in C} exp(dis_{total}(s_{c^{'}}, q)}
\end{equation}
\end{small}

%%%%%%%%%%%%%%%%%%%%%%%%%%%%%%%%%%%%%%%%%%%%%%%%%%%%%%%%%%%%%%%%%%%%%%%%%%%%%%%%%%%%%%
%%%%%%%%%%%%%%%%%%%%%%%%%%%%%%%%%%%%Experiments%%%%%%%%%%%%%%%%%%%%%%%%%%%%%%%%%%%%%%%
%%%%%%%%%%%%%%%%%%%%%%%%%%%%%%%%%%%%%%%%%%%%%%%%%%%%%%%%%%%%%%%%%%%%%%%%%%%%%%%%%%%%%%
\section{Experiments}
\subsection{Datasets}
In our experiments, we use UCF101\cite{soomro2012ucf101}, HMDB51\cite{kuehne2011hmdb}, Kinetics400\cite{carreira2017quo-kinetics}, SSv2-Full (SSv2-Small )\cite{goyal2017something},  . 
The dataset setting of SSv2-small and Kinesics: 
the split methods of HMDB51 and UCF101 follow ARN\cite{zhang2020few-ARN}.
In the UCF101, there are 70 training classes, 10 for validation and 21 for testing, with $9154/1421/2745$ videos for $train/val/test$, respectively. 
The HMDB51 contains 31 classes for training, 10 classes for validation, and 10 classes for testing, with $4280/1194/1292$ videos for $train/val/test$.
we use the split method of CMN\cite{zhu2018compound}, CMN-J\cite{zhu2020label} for the setting of Kinesics and SSv2-small. 
The split method randomly selects a mini-dataset containing 100 classes, including 64 training classes, 12 validation classes, and 24 testing classes, each with 100 samples. 
We follow the OTAM\cite{cao2020few-otam} split method for setting SSv2-Full. This method contains $77500/1926/2854$ videos for $train/val/test$ respectively. 
The class division for SSv2-Full is similar to SSv2-Small, but there are more samples under each class.

\subsection{Implementation Details}
Data augmentation: in the training stage, we flip each frame horizontally and randomly crop the center region 224 × 224.
Backbone: we use both the ResNet50 and VIT-B/16 of CLIP as the visual encoder.
Optimizer: we use the Adam \cite{2014Adam}.
Learning rate: the learning rate is 0.00001.
Video frames: we follow the previous work TSN \cite{wang2016temporal-tsn} for the video frame. Eight frames are sparsely and uniformly sampled from each video.
Training stage: we average gradients and backpropagate once every 16 iterations.
Testing stage: we make 10,000 episodes, and our experiment's average accuracy is reported. For the augmentation, we use only the center crop to augment the video.
We use the OTAM \cite{cao2020few-otam} as the comparison method for distance.

\begin{table*}[ht]
  \centering
  \caption{Comparison on UCF101, HMDB51. Accuracies for 5-way, 3-shot, and 1-shot settings are shown. $\diamond$ means our implementation
  }
  \vspace{-0.5em}
  \textbf{}
    \label{table:1}
    \scalebox{0.8}{
    \begin{tabular}{l|l|ccc|ccc}
    \toprule
    \multicolumn{1}{c|}{\multirow{2}[2]{*}{Method}} & \multicolumn{1}{c|}{\multirow{2}[2]{*}{Backbone}} & \multicolumn{3}{c|}{HMDB51} & \multicolumn{3}{c}{UCF101} \\
               &       & \multicolumn{1}{c}{1-shot} & \multicolumn{1}{c}{3-shot} & \multicolumn{1}{c|}{5-shot} & \multicolumn{1}{c}{1-shot} & \multicolumn{1}{c}{3-shot} & \multicolumn{1}{c}{5-shot} \\
    \midrule
    ProtoNet \cite{snell2017prototypical_few_shot_learning2}&ResNet-50& 54.2&- &   68.4& 70.4& - &89.6 \\
    OTAM\cite{cao2020few-otam}  &ResNet-50&54.5 & - &66.1&79.9&- &88.9\\
    TRX \cite{perrett2021temporal-trx} &ResNet-50& $52.9^{\diamond}$ &- &75.6 &$77.3^{\diamond}$&-&96.1 \\
    STRM \cite{thatipelli2022spatio-strm}&ResNet-50& $54.1^{\diamond}$ &- &77.3&$79.2^{\diamond}$& - &96.9  \\
    HyRSM\cite{wang2022-hybrid}&ResNet-50&60.3  &71.7 & 76.0 & 83.9 & 93.0 & 94.7 \\
    MTFAN\cite{wu2022motion}&ResNet-50& 59.0 &- & 74.6 &84.8 & -& 95.1 \\
    TA2N\cite{li2022ta2n}&ResNet-50& 59.7 & - &73.9 &81.9  & -&95.1  \\
    HCL\cite{zheng2022few-hcl}&ResNet-50& 59.1 & - &76.3 &82.6  & -&94.5  \\
    TADRNet\cite{2023Task-AwareDual-Representation}&ResNet-50&64.3&74.5&78.2&86.7&94.3&96.4 \\
    MoLo\cite{wang2023molo} &ResNet-50& 60.8&72.0&77.4& 86.0&93.5&95.5  \\
    AMeFu-Net\cite{fu2020depth}  &ResNet-50&60.2 &- &75.5& 85.1&-&95.5\\
    SRPN(2021)\cite{wang2021semantic}&ResNet-50 &61.6 & 72.5 &76.2 & 86.5 & 93.8  & 95.8   \\
    TAda-Net\cite{wang2022task}  &ResNet-50&60.8 & 71.8 &76.4  & 85.7 & 93.3 &95.7  \\
    AMFAR\cite{wanyan2023active}  &ResNet-50& 73.9 &- & 87.8& 91.2 &- & 99.0 \\
    CLIP-FSAR\cite{wang2023clip}& CLIP-RN50 &  69.4  & 78.3 & 80.7  &92.4 & 95.4 &97.0\\ 
    CLIP-FSAR\cite{wang2023clip}& CLIP-VIT-B &  77.1  & 84.1 & 87.7  &97.0 & 98.5 & 99.1 \\  
    \midrule
\rowcolor{gray!15}    CLIP-$\mathrm{M^2}$DF      &CLIP-RN50& 66.8 & 79.4 $\uparrow$  & 83.0$\uparrow$ & 94.3 $\uparrow$ & 98.0$\uparrow$& 98.8$\uparrow$  \\
\rowcolor{gray!15}    CLIP-$\mathrm{M^2}$DF      &CLIP-VIT-B& 77.0  & 84.1 $\uparrow$ & 88.0$\uparrow$ &  97.0 &  98.1     & 99.3$\uparrow$  \\
    \bottomrule
    \end{tabular}%
    }
\vspace{-0.5em}
\end{table*}%

%%%%%%%%%%%%%%%%%%%%%%%%%%%%%%%%%%%%%%%%%%%%%%%%%%%%%%%%%%%%%%%%%%%%%%%%%%%%%%%%%%%%%

\begin{table*}[ht]
  \centering
  \caption{Comparison on Kinetics and SSv2 datasets. Accuracy results for 5-way, 3-shot, and 1-shot settings are shown. $\diamond$ means the results of our implementation (the data in parentheses represents the published data).}
  \vspace{-0.5em}
    \label{table:2}
    \scalebox{0.8}{
    \begin{tabular}{c|c|ccc|ccc|ccc}
    \toprule
    \multicolumn{1}{c|}{\multirow{2}[2]{*}{Method}} & \multicolumn{1}{c|}{\multirow{2}[2]{*}{Backbone}} & \multicolumn{3}{c|}{Kinetics} & \multicolumn{3}{c|}{SSv2-Full} & \multicolumn{3}{c}{SSv2-small} \\
          &       & \multicolumn{1}{c}{1-shot} & \multicolumn{1}{c}{3-shot} & \multicolumn{1}{c|}{5-shot} & \multicolumn{1}{c}{1-shot} & \multicolumn{1}{c}{3-shot} & \multicolumn{1}{c|}{5-shot} & \multicolumn{1}{c}{1-shot} & \multicolumn{1}{c}{3-shot} & \multicolumn{1}{c}{5-shot} \\
    \midrule
    ProtoNet \cite{snell2017prototypical_few_shot_learning2}& ResNet-50& 65.4 &-&77.9&-& &-&33.6 &- & 43.0 \\
    Matching Net\cite{vinyals2016matching—few_shot_learning5}&ResNet-50 &53.3& -& 74.6 &-&-&-& 34.4 & -&43.8 \\
    OTAM\cite{cao2020few-otam}          &ResNet-50&73.0 &- &85.5 &42.8 &- & 52.3 &$38.9^{\diamond}$ &- & $48.1^{\diamond}$ \\
    TRX \cite{perrett2021temporal-trx}  &ResNet-50& $63.4^{\diamond}$(63.6)  &- & $85.1^{\diamond}$(85.9) &42.0 &-& $63.0^{\diamond}$(64.6)  &36.0 &-& $56.3^{\diamond}$(59.4) \\
    STRM \cite{thatipelli2022spatio-strm}&ResNet-50& $65.3^{\diamond}$ & -  &$85.9^{\diamond}$(86.7)  &$42.9^{\diamond}$ &- &$64.8^{\diamond}$(68.1) &  &- &   \\
    HyRSM\cite{wang2022-hybrid}         &ResNet-50& 73.7  &83.5 & 86.1 &54.3  & 65.1  &  69.0 & 40.6 & 52.3  &56.1  \\
    MTFAN\cite{wu2022motion}           &ResNet-50& 74.6  &- & 87.4 &45.7 &  -  &  60.4 & - & -  &-  \\
    TA2N\cite{li2022ta2n}               &ResNet-50&72.8& -& 85.8 & 47.6 & -&61.0  & - & - & -   \\
    HCL\cite{zheng2022few-hcl}              &ResNet-50&73.7& -& 85.8 & 47.3 & -&64.9  &38.7& - & 55.4   \\
    TADRNet\cite{2023Task-AwareDual-Representation}&ResNet-50& 75.6 & 84.8 & 87.4  &43.0 &-  &61.1&- & - &- \\
    MoLo\cite{wang2023molo}             &ResNet-50&74.0 &83.7& 85.6 &56.6 &67.0& 70.6 & 42.7 &52.9& 56.4 \\
    AMeFu-Net\cite{fu2020depth}  &ResNet-50& 74.1  &- &86.8 &- &-&- & - &-& - \\
    CMN++\cite{zhu2020label}      &ResNet-50&60.5 &75.6 &78.9 & 36.2& 44.6 & 44.8 & - &-  &-  \\
    SRPN\cite{wang2021semantic}   &ResNet-50&75.2 & 84.7 &87.1  & -   & -   & -   & -  &  -  & - \\
    AMFAR\cite{wanyan2023active}  &ResNet-50&80.1 & -& 92.6 &61.7& -& 79.5 & -  & -  &-   \\
    CLIP-FSAR\cite{wang2023clip}  &CLIP-RN50 &90.1&90.8&$91.6^{\diamond}$(92.0)& 58.7&60.7&$62.9^{\diamond}$(62.8) & 52.1 &54.0&$55.3^{\diamond}$(55.8)\\ 
    CLIP-FSAR\cite{wang2023clip}  &CLIP-VIT-B &94.8&95.0&95.4& 62.1&68.3&72.1 & 54.6 &59.4&61.8\\ 
    \midrule
    \rowcolor{gray!15}    CLIP-$\mathrm{M^2}$DF       &CLIP-RN50& 90.1 & 92.0 $\uparrow$ & 93.5$\uparrow$ & 56.9& 63.9 $\uparrow$ & 68.0 $\uparrow$&   51.6 & 56.3 $\uparrow$ & 60.4 $\uparrow$  \\
    \rowcolor{gray!15}   CLIP-$\mathrm{M^2}$DF        &CLIP-VIT-B&90.9 & 95.1 $\uparrow$ & 96.2 $\uparrow$ & 60.1 &  68.9 $\uparrow$ & 72.7 $\uparrow$ & 52.0 &  58.9 &  62.8 $\uparrow$ \\
    \bottomrule
    \end{tabular}

    }
\vspace{-0.5em}
\end{table*}

\subsection{Comparison with the-state-of-the-art works}
We compare our model with the state-of-the-art methods, and our baseline is the CLIP-FSAR.
Firstly, we use CLIP-RN50 as the backbone.
For the UCF101 and HMDB51, see in the \cref{table:1}:
in the 5-shot setting,
our CLIP-$M^2$DF is significantly superior to CLIP-FSAR with $1.8\%$ and $2.3\%$. 
In the 3-shot setting, our CLIP-$M^2$DF is superior to CLIP-FSAR from $95.4\%$ to $98.0\%$ and from $78.3\%$ to $79.4\%$. 
For the Kinetics, SSv2-Full and SSv2-Small, see in the \cref{table:2}:
in the 5-shot setting, 
our CLIP-$M^2$DF is still superior to CLIP-FSAR with $1.5\%$, $5.2\%$ and $4.6\%$.
In the 3-shot setting, our CLIP-$M^2$DF is superior to CLIP-FSAR from $90.8\%$ to $92.0\%$, from $60.7\%$ to $63.9\%$  and from $54.0\%$ to $56.3\%$. 
With CLIP-RN50, compared to AMFAR, the accuracies under HDMB51 and SSV-Full are lower. However, AMFAR introduces optical flow data, which is sometimes difficult to obtain in the production environment.
We also give the results based on CLIP-VIT. 
In SSv2-Full, the AMFAR is better than ours because SSv2 is a time-sensitive dataset. The increased optical flow information has made a significant contribution.
For other datasets, our model gets the state-of-the-art result except for the 1-shot setting. 
According to the \cref{table:1} and \cref{table:2}, we can conclude that: 
(1) Our model is better than most of the current methods. 
(2) Our model is better than CLIP-FSAR. 
(3) Our model with CLIP-RN50 is competitive with AMFAR. Except for SSv-Full, Our model with CLIP-VIT is better than AMFAR even though optical flow data are used in AMFAR.
%%%%%%%%%%%%%%%%%%%%%%%%%%%%%%%%%%%%%%%%%%%%%%%%%%%%%%%%%%%%%%
%%%%%%%%%%%%%%%%%%%%%%%%%%%%%%%%%%%%%%%%%%%%%%%%%%%%%%%%%%%%%%
\subsection{Ablation Study}
To prove all the parts of our model are effective, we design several experiments.
Firstly, we design an experiment to verify Single-view and PPS. 
Secondly, we experiment to verify the Multi-view fusion of bidirectional distillation under both distillation conditions and PPS.
Thirdly, an experiment is designed for the distillation condition and direction.
Our ablation study is based on CLIP-RN50. 

\subsubsection{Probability Prompt Selector and Single-view Ablation}
To prove that the view of LTCE (GTCE) with MMFE and PPS is effective, 
we experiment to compare the views of GTCE, LTCE, and NTCE. 
The baseline is NTCE (None Temporal Context Extractor), which is a Transformer using the same $Query$, $Key$, and $Value$ that are not processed by the Temporal Context Extractor. 
In the \cref{table:3}, no matter if we use LTCE, GTCE, or NTCE view, in most of the settings, the accuracy with the PPS is higher than without it. That means the probability prompt for the query could supplement the class consistency information.
Also, no matter if we use the PPS, the accuracy of both the GTCE view and LTCE view are better than the NTCE view, which means they are effective.

\begin{table}[htbp]
  \centering
  \caption{Compare the Single-view with/without Temporal Context Extractor and PPS}
  \vspace{-0.5em}
  \scalebox{0.7}{
    \begin{tabular}{c|c|c|c|c|c|c|c}
    \toprule
    Single-view& \multicolumn{3}{c|}{Single-view} & \multirow{2}[2]{*}{PPS} & \multicolumn{1}{c|}{Kinetics} & \multicolumn{1}{c|}{SSv2-Small} &\multicolumn{1}{c}{HMDB51} \\
   Serial No. &  \multicolumn{1}{c}{GTCE}&  \multicolumn{1}{c}{LTCE} &  \multicolumn{1}{c|}{NTCE} &      & \multicolumn{1}{c|}{5-shot} & \multicolumn{1}{c|}{5-shot} & \multicolumn{1}{l}{5-shot} \\
    \midrule
    1&\ding{51}&\ding{55}&\ding{55}&\ding{55}&92.0& 58.1& 82.4 \\
    2&\ding{51}&\ding{55}&\ding{55}&\ding{51}&93.2& 58.5& 82.4 \\
    3&\ding{55}&\ding{51}&\ding{55}&\ding{55}&91.9& 56.2& 82.1 \\
    4&\ding{55}&\ding{51}&\ding{55}&\ding{51}&92.9& 57.2& 82.3\\
    5&\ding{55}&\ding{55}&\ding{51}&\ding{55}&91.4& 56.0& 81.7 \\
    6&\ding{55}&\ding{55}&\ding{51}&\ding{51}&92.0& 56.5& 82.2 \\
    \bottomrule
    \end{tabular}%
    }
  \label{table:3}
\vspace{-0.8em}
\end{table}%

%%%%%%%%%%%%%%%%%%%%%%%%%%%%%%%%%%%%%%%%%%%%%%%%%%%%%%%%%%%%%%%%%%
\subsubsection{Ablation for Multi-view fusion with Distillation and PPS}
In the \cref{table:4}, using distillation and PPS achieves the highest accuracy. The model with distillation is better than without it. The same result is for the PPS.
When the item is without PPS, the Multi-view distillation effect is limited. For Kinetics, the accuracy is from $92.2\%$ to $92.3\%$, for SSv2-Small from $56.6\%$ to $57.1\%$, and for HMDB51 from $82.1\%$ to $82.4\%$. 
When we add PPS, the distillation effect increases. For Kinetics, the accuracy is from $93.1\%$ to $93.5\%$, for SSv2-Small, from $59.1\%$ to $60.4\%$, and for HMDB51 from $82.5\%$ to $83.0\%$.
For the items in \cref{table:4} compared with the according items in \cref{table:3}, we want the accuracy of Multi-view to be higher than the related Single-view.
But for SSv2-Small,
we can see the Multi-view fusion No.3 without PPS and distillation ($56.6\%$) and No.1 just without PPS ($57.1\%$) are lower than the Single-view No.1 GTCE without PPS ($58.1\%$).
For the HMDB51, Multi-view No.1 just without PPS ($82.4\%$) is equal to the Single-view No.1 GTCE without PPS ($82.4\%$).
Perhaps the reason is that we only simply fuse the distances of the Multi-view, 
see in \cref{eq9}, and do not add any restrictions so that the fusion distance or distillation may not always increase discriminative ability.
Perhaps fusion and distillation can produce greater reactions based on better data materials.

\begin{table}[htbp]
  \centering
  \caption{Multi-view Distillation and PPS}
  \vspace{-0.5em}
  \scalebox{0.7}{
    \begin{tabular}{c|c|c|c|c|c}
    \toprule
    Multi-view fusion  & \multirow{2}[2]{*}{Distillation} & \multirow{2}[2]{*}{PPS} & \multicolumn{1}{c|}{Kinetics} & \multicolumn{1}{c|}{SSv2-Small} & \multicolumn{1}{c}{HMDB51} \\
    Serial No. &       &       & \multicolumn{1}{c|}{5-shot} & \multicolumn{1}{c|}{5-shot} & \multicolumn{1}{c}{5-shot} \\
    \midrule
      1&\ding{51}& \ding{55}&92.3   & 57.1  & 82.4 \\
      2&\ding{51}& \ding{51}&93.5   & 60.4  & 83.0 \\
      3&\ding{55}& \ding{55}&92.2   & 56.6 &  82.1 \\
      4&\ding{55}& \ding{51}&93.1   & 59.1 &  82.5 \\
    \bottomrule
    \end{tabular}%
    }
  \label{table:4}
  \vspace{-0.8 em}
\end{table}%

%%%%%%%%%%%%%%%%%%%%%%%%%%%%%%%%%%%%%%%%%%%%%%%%%%%%%%%%%%%%%%%%%%%%%
\subsubsection{Multi-view Distillation Condition and Direction}
In the \cref{table:5}, v-compare means visual comparison condition, and t-compare means token comparison condition. Up and down mean the LTC view and GTC view.
While using both the t-compare and v-compare, together with the bidirectional mutual distillation, our model can achieve the highest accuracy for Kinetics, SSv2-Small, and HMDB51. 
Using both distillation conditions, the accuracy is higher than just using one.
From an overall view, the accuracy of bidirectional distillation is higher than that of unidirectional distillation. 

\begin{table}[htbp]
  \centering
  \caption{Distill condition and direction study.}
  \vspace{-0.5em}
  \scalebox{0.7}{
    \begin{tabular}{c|c|c|c|c|c}
    \toprule
    \multicolumn{2}{c|}{Distillation condition} & \multirow{2}[2]{*}{Distillation Direction} & \multicolumn{1}{c|}{Kinetics} & \multicolumn{1}{c|}{SSv2-Small} & \multicolumn{1}{c}{HMDB51} \\
    \multicolumn{1}{c}{t-compare} & v-compare   &       & \multicolumn{1}{c|}{5-shot} & \multicolumn{1}{c|}{5-shot} & \multicolumn{1}{l}{5-shot} \\
    \midrule
    \ding{51}   & \ding{51}    & bidirectional  &93.5 &60.4 & 83.0 \\
    \ding{51}   & \ding{55}    & bidirectional  & 93.0 & 59.2 &82.6 \\
    \ding{55}   & \ding{51}    & bidirectional  & 93.1 &59.7 & 82.3\\
    \ding{55}   & \ding{55}    & bidirectional  & 93.0 & 58.9 &82.3\\
    \ding{51}   & \ding{51}    & up$\rightarrow$down & 93.2&59.1 &82.6\\
    \ding{51}   & \ding{55}    & up$\rightarrow$down & 93.1&58.8 &82.3\\
    \ding{55}   & \ding{51}    & up$\rightarrow$down &93.0 & 58.9 &82.4\\
    \ding{55}   & \ding{55}    & up$\rightarrow$down &92.7& 58.4 & 82.2\\
    \ding{51}   & \ding{51}    & down$\rightarrow$up &93.1&58.9&82.5\\
    \ding{51}   & \ding{55}    & down$\rightarrow$up &93.1& 58.0&82.3\\    
    \ding{55}   & \ding{51}    & down$\rightarrow$up &92.9& 59.3&82.1\\
    \ding{55}   & \ding{55}    & down$\rightarrow$up &92.7&58.6&82.1\\
    \bottomrule
    \end{tabular}%
  }
  \label{table:5}
\vspace{-0.5em}
\end{table}%

\subsubsection{Analysis of Distillation Hyper-Parameter}
\begin{figure}[ht] 
\centering
\includegraphics[width=7.3cm]{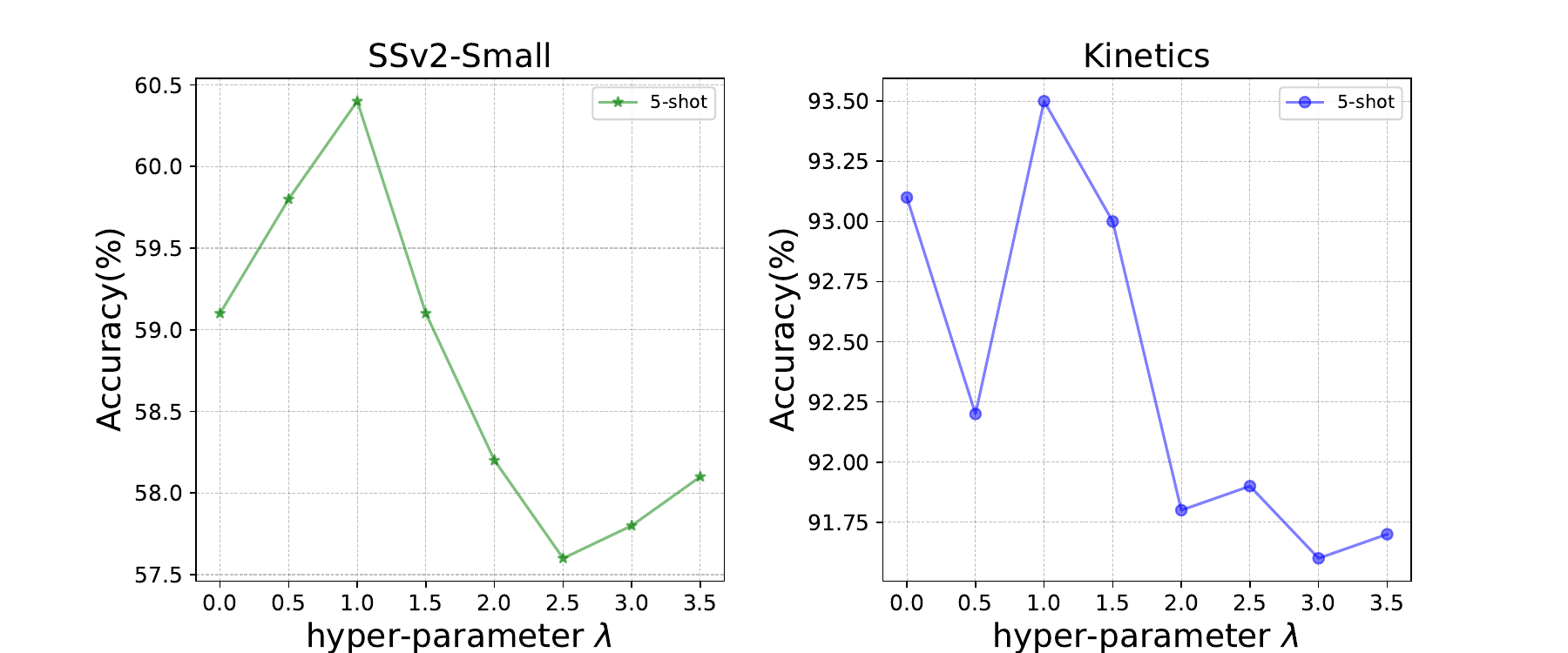}
\caption{ Hyper-parameter $\lambda$ for accuracy.
} 
\label{fig:4-1}
\vspace{-0.8em}
\end{figure}
In the \cref{eq15}, a hyper-parameter is used to balance the contribution of the distillation loss and distance loss. In \cref{fig:4-1}, the hyper-parameter has a certain impact on the accuracy of few-shot recognition. We can see the recognition accuracy is highest near 1 both for SSV2-Small and Kinetics. And in all the other experiments, we keep the hyper-parameter to be 1.
%%%%%%%%%%%%%%%%%%%%%%%%%%%%%%%%%%%%%%%%%%%%%%%%%%%%%%%%%%%%%%%%%%%%%%%%%%%%%%%%%%%%%%
%%%%%%%%%%%%%%%%%%%%%%%%%%%%%%%%%%%%%Visual%%%%%%%%%%%%%%%%%%%%%%%%%%%%%%%%%%%%%%%%%%
\section{Visual}
We have demonstrated a series of visualizations to prove the progressiveness of our model.
\subsection{Distribution comparison}
To demonstrate that labels and temporal context bring class consistency information in each context view, we use t-SNE \cite {wang2016temporal-tsn} to plot the data distribution for Kinetics and SSv2 under 5-way 5-shot settings.
\begin{figure}[t]
    \centering
    \subfloat[CLIP-FSAR]{
    \label{fig:5.a}
    \includegraphics[width=3cm]{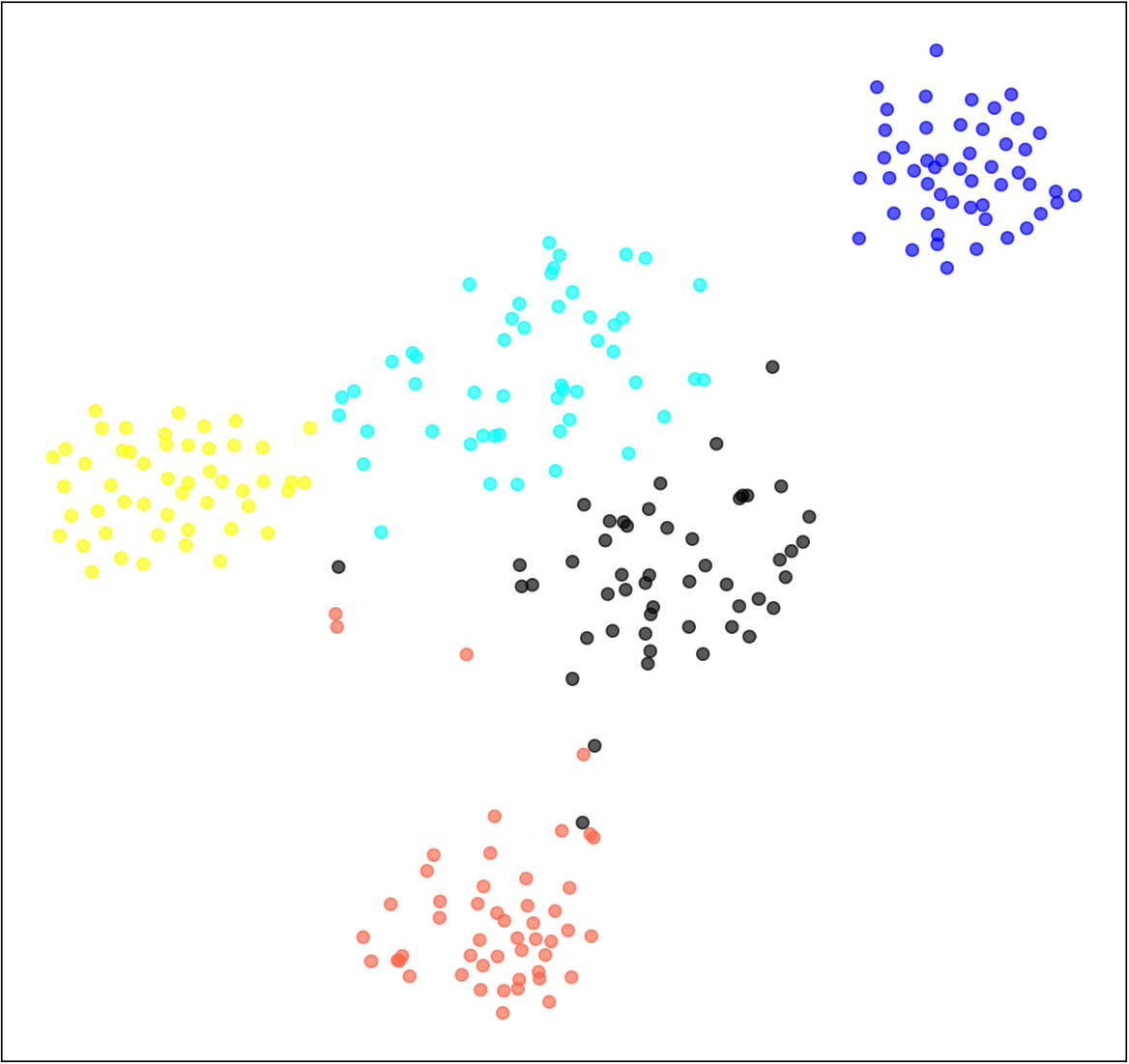}
    }
    \subfloat[Ours-LTC View]{
    \label{fig:5.b}
    \includegraphics[width=3cm]{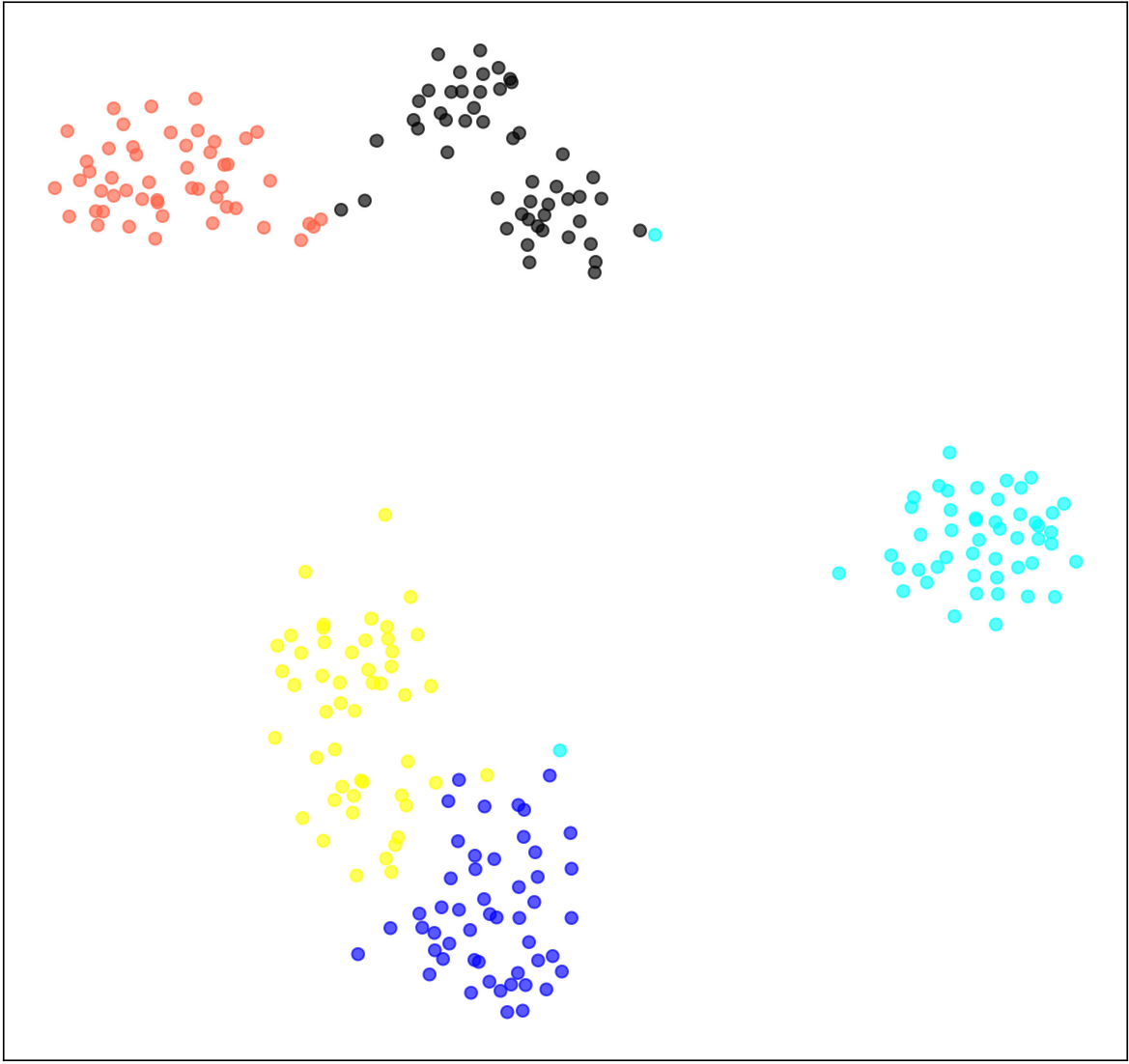}
    }
        \subfloat[Ours-GTC View]{
    \label{fig:5.c}
    \includegraphics[width=3cm]{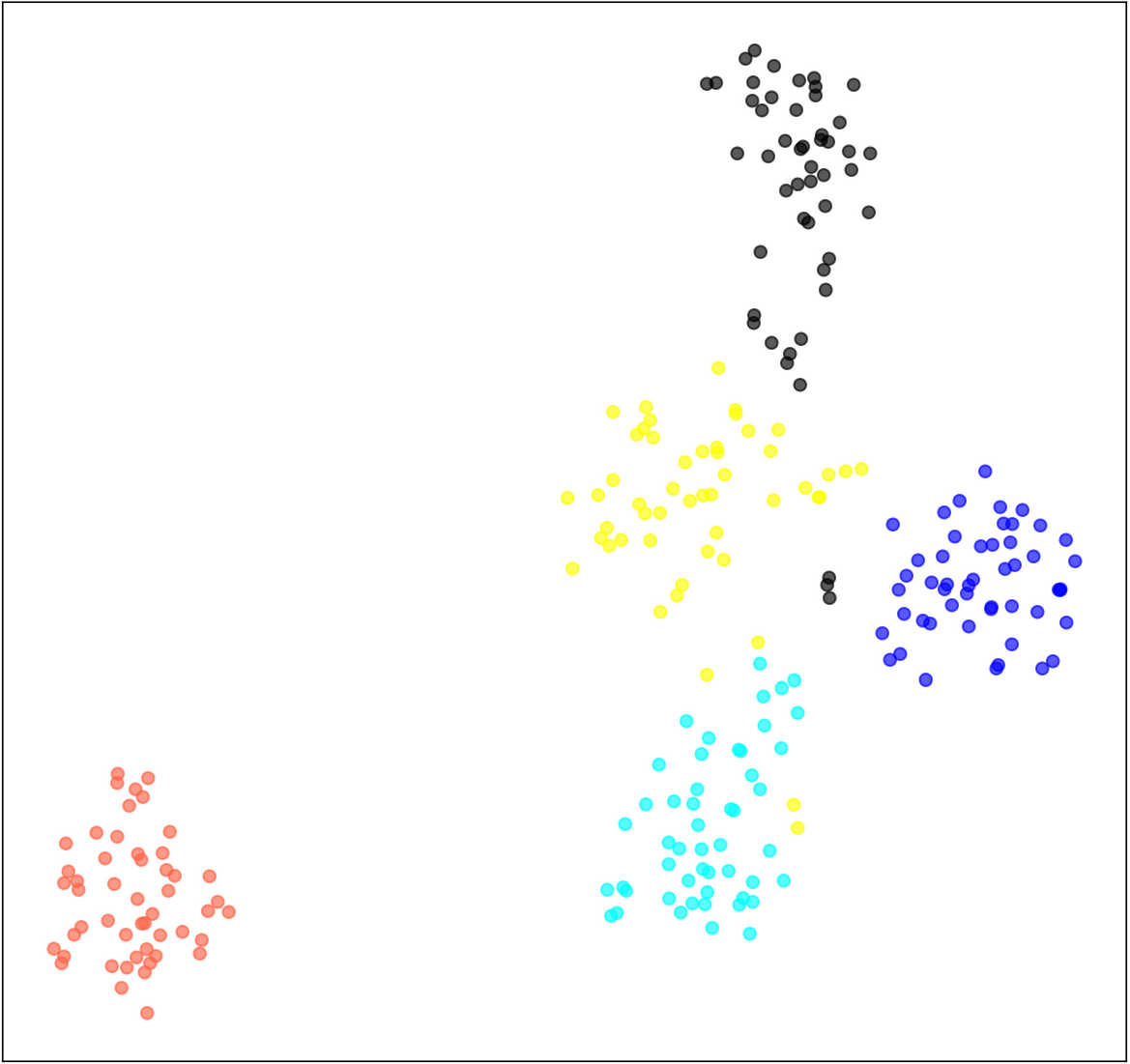}
    }
    \caption{Distribution Comparison on Kinetics.
    }
\vspace{-0.8em}
\label{fig:5}
\end{figure}

\begin{figure}[htbp]
    \centering
    \subfloat[CLIP-FSAR]{
    \label{fig:6.a}
    \includegraphics[width=3cm]{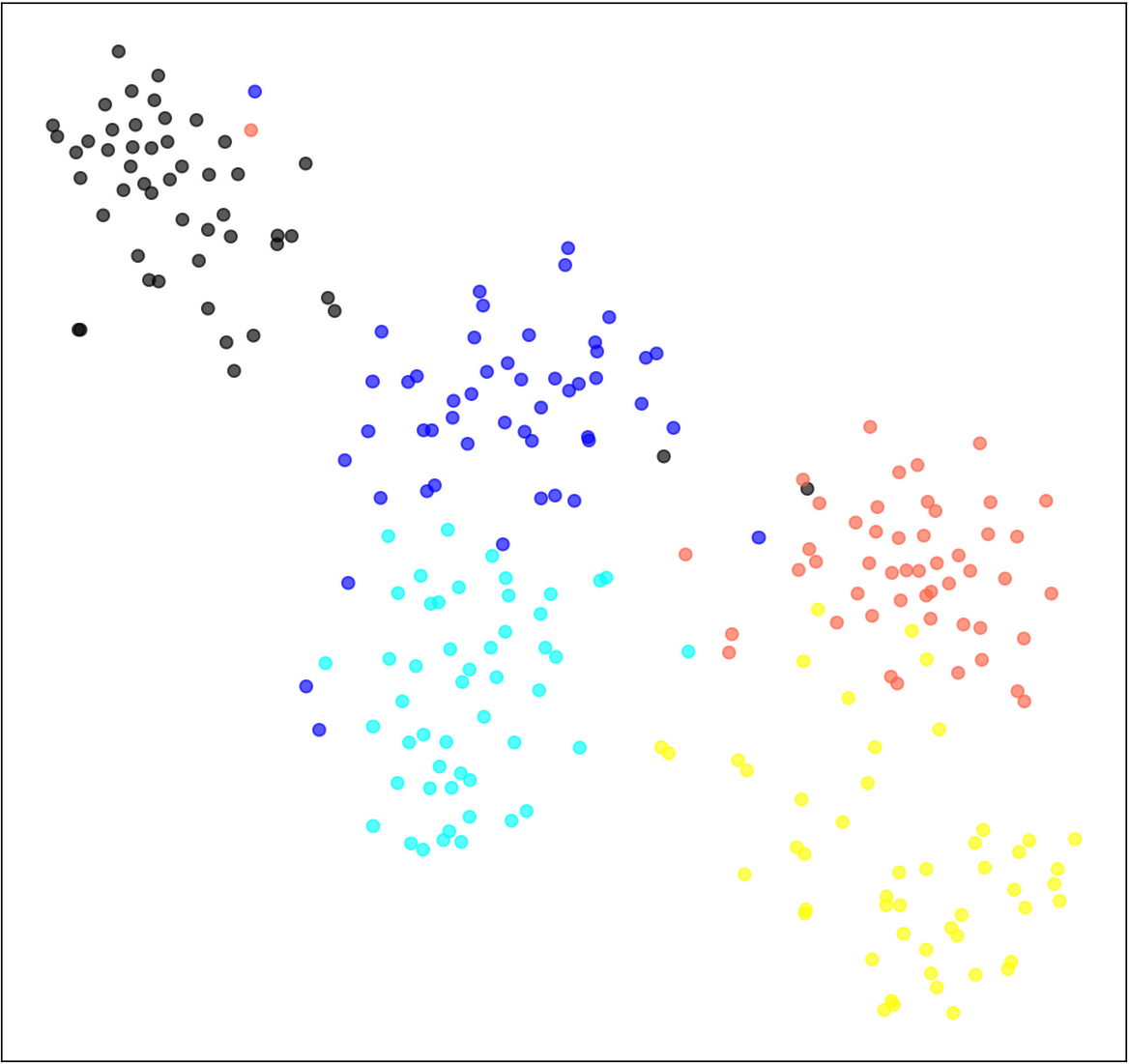}
    }
    \subfloat[Ours-LTC View]{
    \label{fig:6.b}
    \includegraphics[width=3cm]{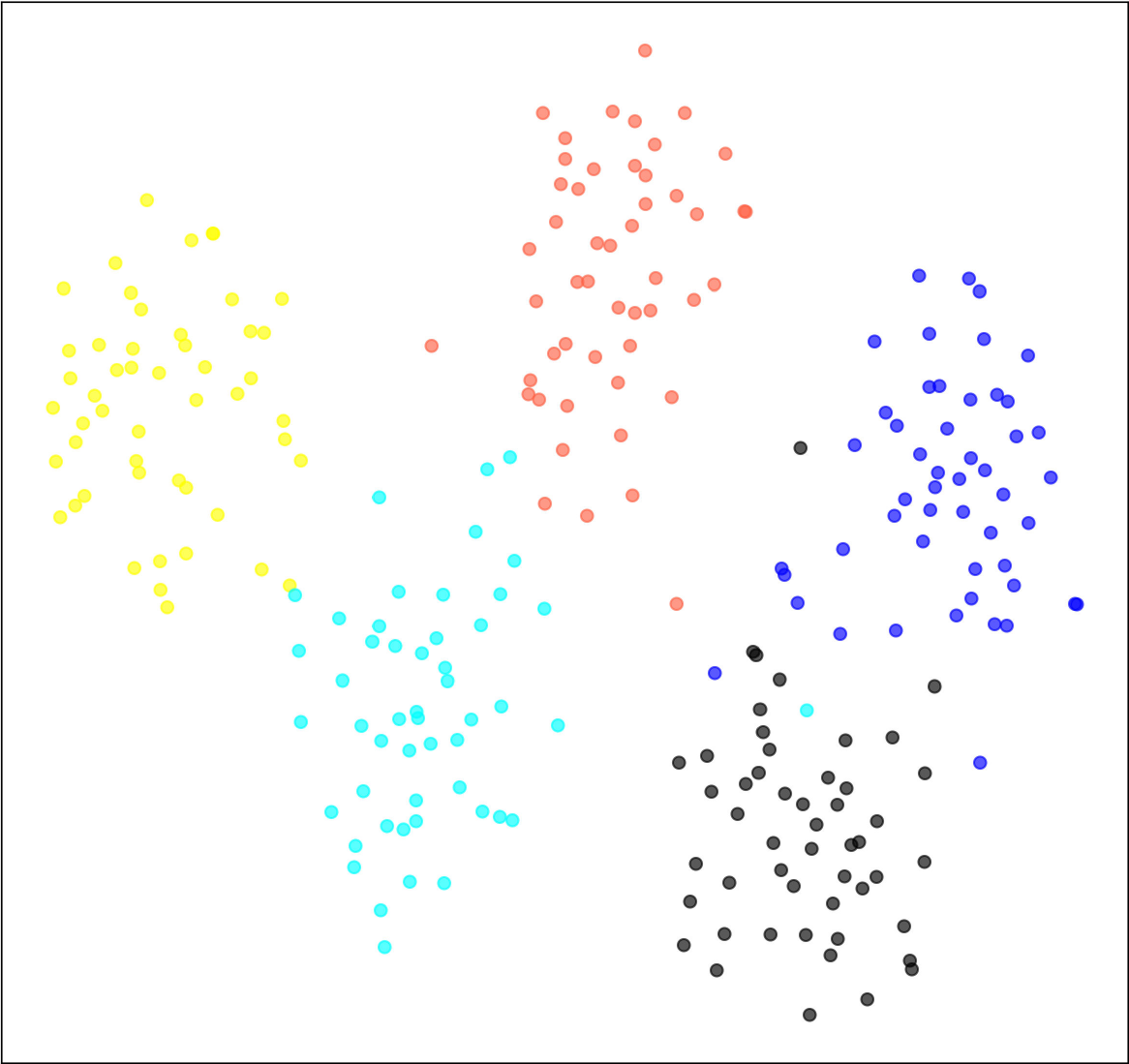}
    }
        \subfloat[Ours-GTC View]{
    \label{fig:6.c}
    \includegraphics[width=3cm]{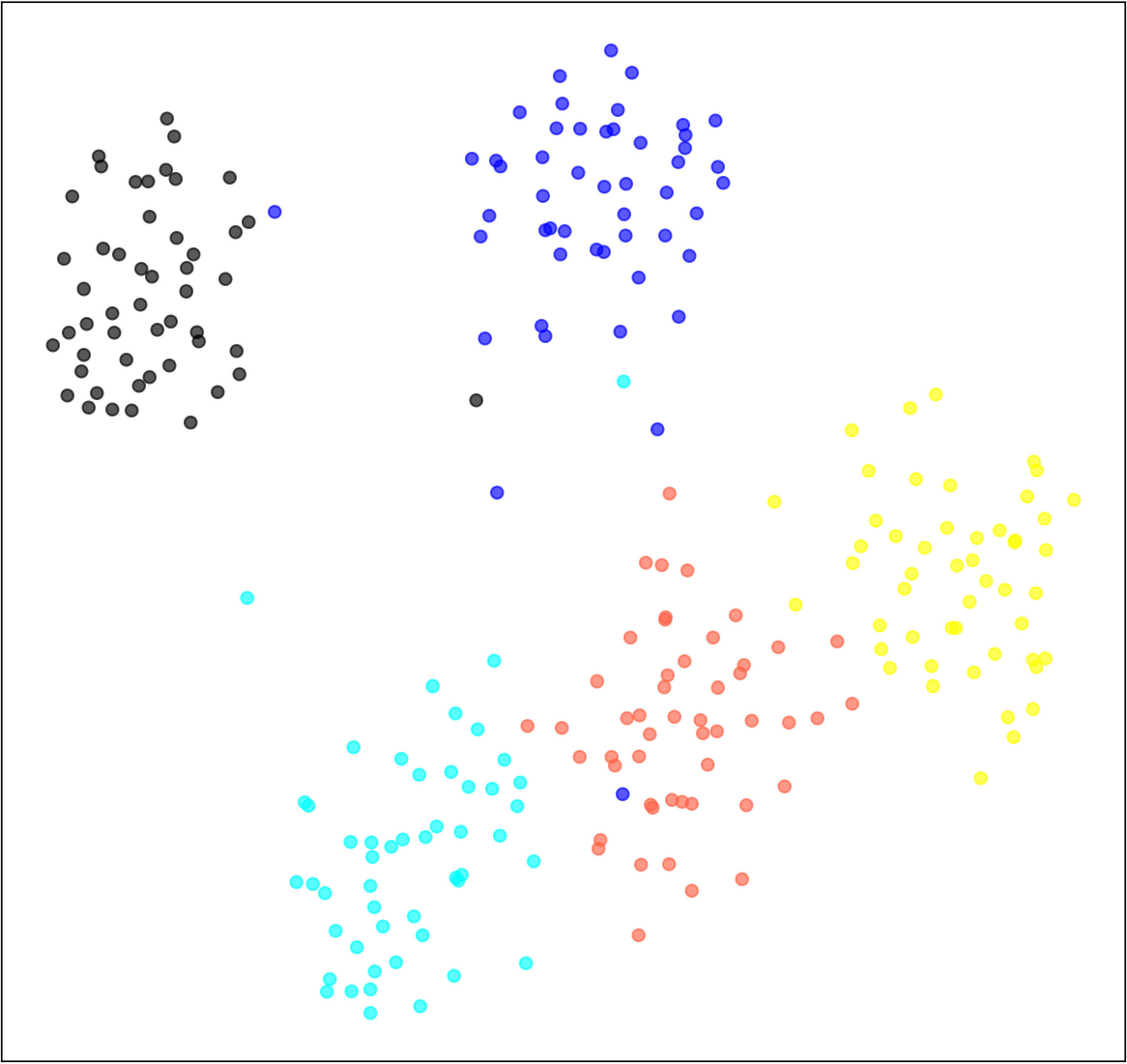}
    }
    \caption{Distribution Comparison on SSv2.
    }
\vspace{-0.8em}
\label{fig:6}
\end{figure}

In the \cref {fig:5} and \cref {fig:6}, The data distribution of CLIP-FSAR and the Single-view of our model that contains visual information, text information, and temporal context information have been shown for Kinetics and SSv2-Small. 
These figures, regardless of the LTC View or the GTC View, demonstrate that inter-class distribution is more discriminative with text information and temporal context information, and intra-class distribution is more stable. Also, there are fewer outliers in the views of our model than CLIP-FSAR. 

%%%%%%%%%%%%%%%%%%%%%%%%%%%%%%%%%%%%%%%%%%%%%%%%%%%%%%%%%%%%%%%%%%
\subsection{Accuracy comparison of different classes}

\begin{figure}[t]
    \centering
    \subfloat[Accuracy under Kinetics]{
    \label{fig:7.a}
    \includegraphics[width=7.2cm]{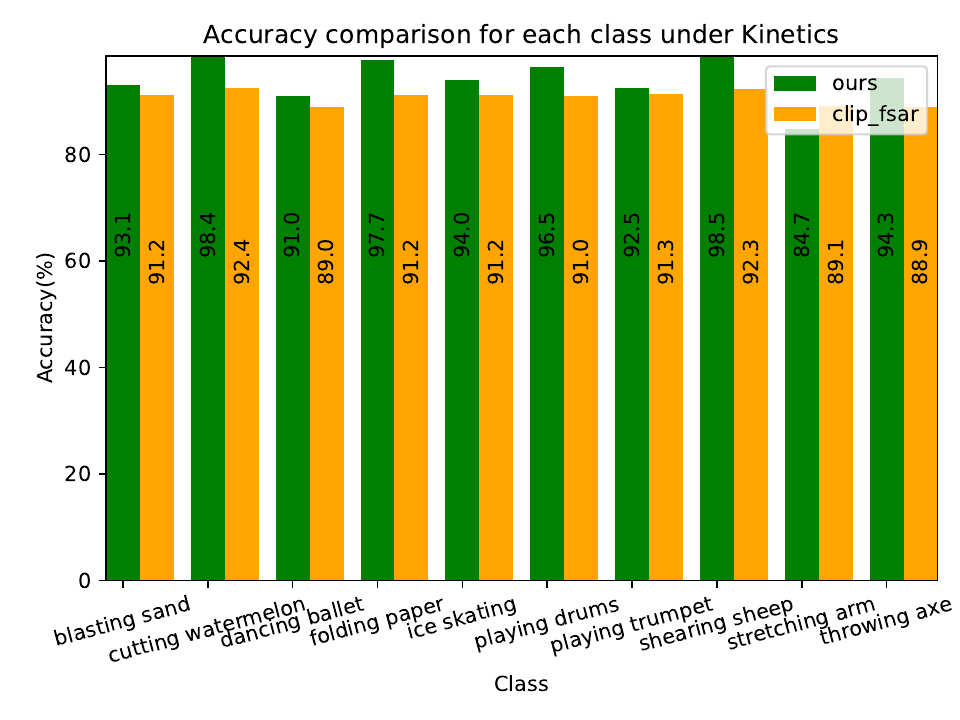}
    }
    \\
    \subfloat[Accuracy under SSv2-small]{
    \label{fig:7.b}
    \includegraphics[width=7.2cm]{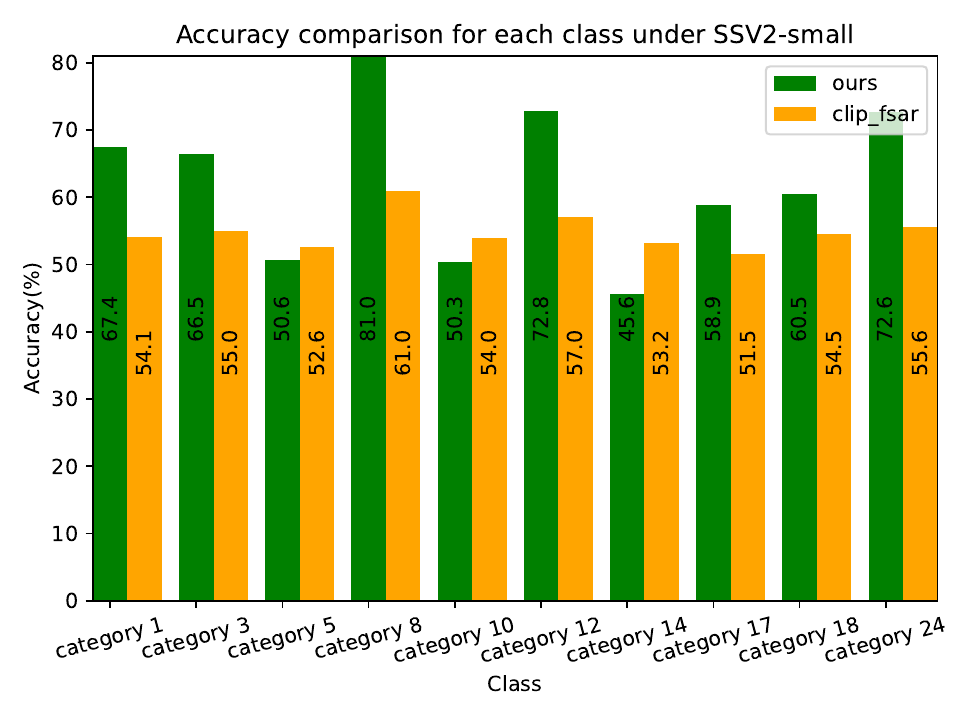}
    }
    \caption{The illustration shows the comparison.
    }
\label{fig:7}
\vspace{-0.8em}
\end{figure}
Compared with CLIP-FSAR \cite{wang2023clip} on Kinetics, 10 classes were randomly selected from 24 classes in the meta-testing stage under the 5-shot setting. 
The results are illustrated in \cref{fig:7.a} and \cref{fig:7.b}, reveal notable accuracy improvements for CLIP-$\mathrm{M^2}$DF across various classes. 
Notably, in Kinetics, the accuracy for the "shearing sheep" class demonstrates the most improvement, and the accuracy of CLIP-$\mathrm{M^2}$DF is $98.5\%$,
and the accuracy of CLIP-FSAR is $92.3\%$.
Similarly, in SSv2-small, for most of the classes, the accuracy of our model is higher than that of CLIP-FSAR. 
For the class "Poking a stack of something so the stack collapses", the accuracy increases significantly from $61.0\%$ of CLIP-FSAR to $81.0\%$ of CLIP-$\mathrm{M^2}$DF. These results demonstrate that CLIP-$\mathrm{M^2}$DF effectively enhances the accuracy of specific action classes.

Note: The class labels we select for SSV2-small are as follows:
``Dropping something into something", 
``Letting something roll up a slanted surface, so it rolls back down",
``Opening something", 
``Poking a stack of something so the stack collapses", 
``Pushing something off of something", 
``Putting something next to something", 
``Putting something on the edge of something so it is not supported and falls down", 
``Scooping something up with something", 
``something falling like a feather or paper', 
``Unfolding something".

\subsection{Comparison for Attention visualization}
\begin{figure}[htbp]
    \centering
    \subfloat[Origin Frames]{
    \label{fig:8.a}
    \includegraphics[width=7.2cm]{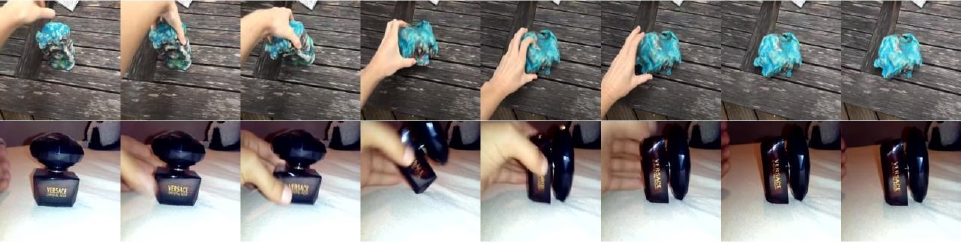}
    }
    \\
    \subfloat[CLIP-FSAR (RN50)]{
    \label{fig:8.b}
    \includegraphics[width=7.2cm]{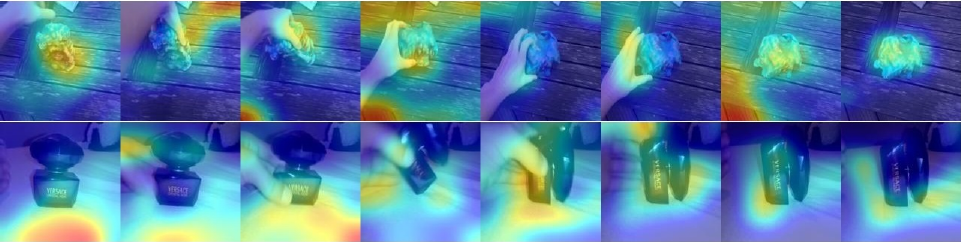}
    }
    \\
    \subfloat[CLIP-$\mathrm{M^2}$DF (RN50)]{
    \label{fig:8.c}
    \includegraphics[width=7.2cm]{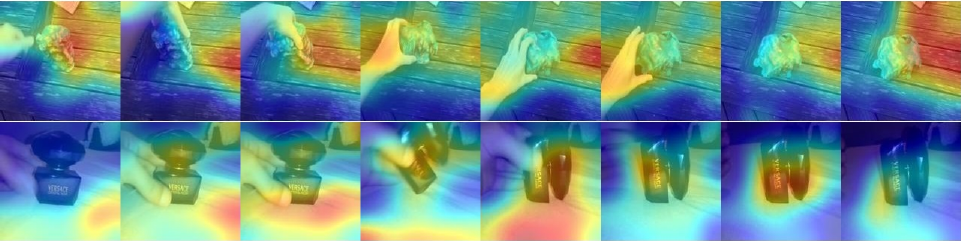}
    }
    \label{fig:attention_map_ssv_res}
    \caption{Attention visualization for SSv2 based on CLIP-RN50 in 5-way 5-shot setting.}
\label{fig:8}
\end{figure}

\begin{figure}[htbp]
    \centering
    \subfloat[Origin Frames]{
    \label{fig:9.a}
    \includegraphics[width=7.2cm]{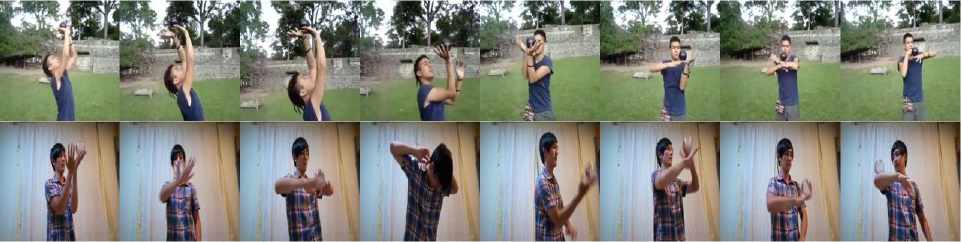}
    }
    \\
    \subfloat[CLIP-FSAR (RN50)]{
    \label{fig:9.b}
    \includegraphics[width=7.2cm]{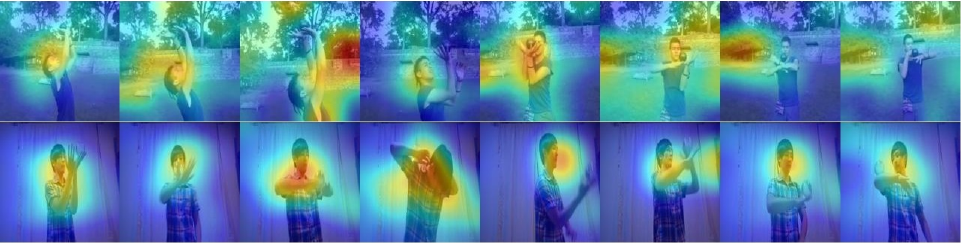}
    }
    \\
    \subfloat[CLIP-$\mathrm{M^2}$DF (RN50)]{
    \label{fig:9.c}
    \includegraphics[width=7.2cm]{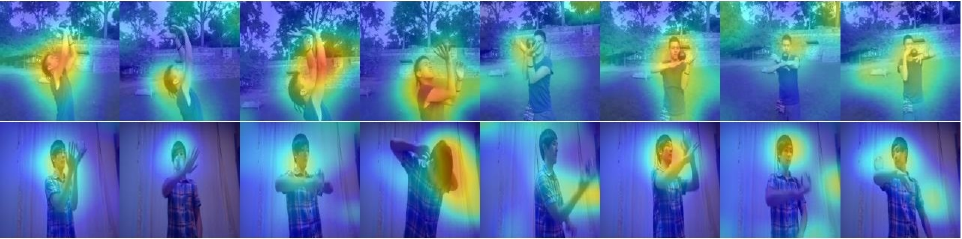}
    }
    \label{fig:attention_map_kinetics_res}
    \caption{Attention visualization for Kinetics based on CLIP-RN50 in 5-way 5-shot setting.}
\label{fig:9}
\vspace{-0.8em}
\end{figure}

\begin{figure}[t]
    \centering
    \subfloat[Origin Frames]{
    \label{fig:10.a}
    \includegraphics[width=7.2cm]{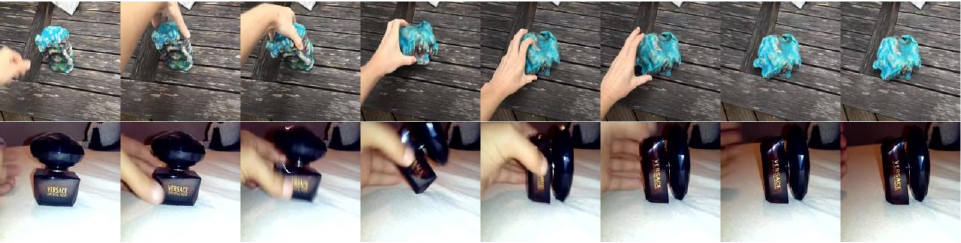}
    }
    \\
    \subfloat[CLIP-FSAR (VIT-B/16)]{
    \label{fig:10.b}
    \includegraphics[width=7.2cm]{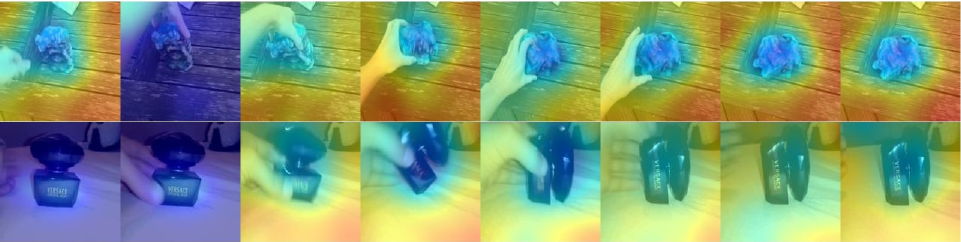}
    }
    \\
    \subfloat[CLIP-$\mathrm{M^2}$DF (VIT-B/16)]{
    \label{fig:10.c}
    \includegraphics[width=7.2cm]{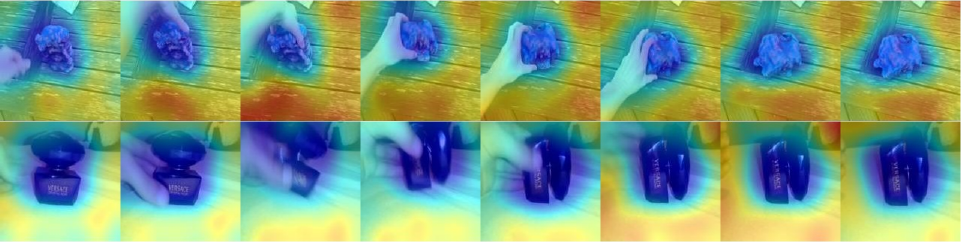}
    }
    \label{fig:attention_map_ssv2_vit}
    \caption{Attention visualization for SSv2 based on VIT-B/16 in 5-way 5-shot setting.}
\label{fig:10}
\vspace{-0.8em}
\end{figure}

\begin{figure}[htpb]
  
    \centering
    \subfloat[Origin Frames]{
    \label{fig:11.a}
    \includegraphics[width=7.2cm]{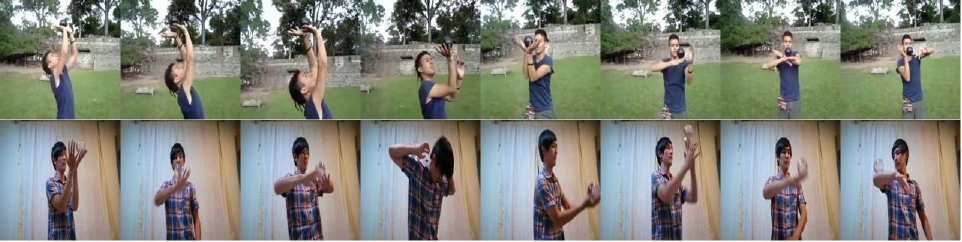}
    }
    \\
    \subfloat[CLIP-FSAR (VIT-B/16)]{
    \label{fig:11.b}
    \includegraphics[width=7.2cm]{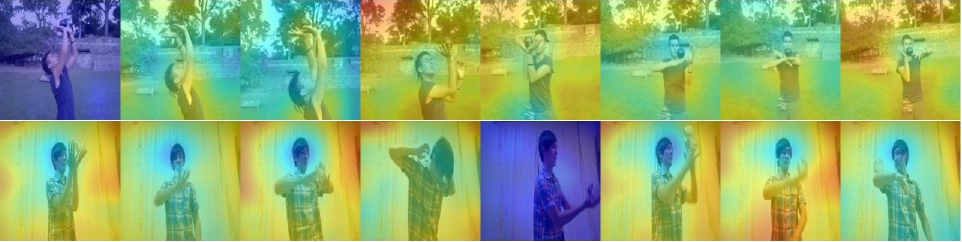}
    }
    \\
    \subfloat[CLIP-$\mathrm{M^2}$DF (VIT-B/16)]{
    \label{fig:11.c}
    \includegraphics[width=7.2cm]{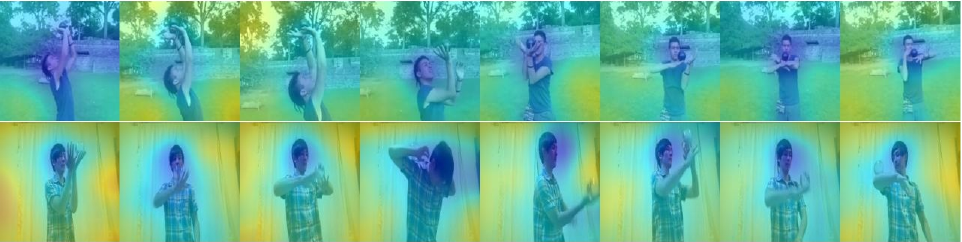}
    }
    \label{fig:attention_map_kinetics_vit}
    \caption{Attention visualization for Kinetics based on VIT-B/16 in 5-way 5-shot setting.}
\label{fig:11}
\vspace{-0.8em}
\end{figure}

To further study the features, attention visualizations of our model are performed and compared with the attention visualizations of CLIP-FSAR. We use the RN50 and VIT-B/16 as our backbone. From \cref{fig:8} to \cref{fig:11}, in each figure, according to the RGB image sequence in sub-figure (a), the attention visualizations of CLIP-FSAR in sub-figure (b) are compared with the attention visualizations of our model in sub-figure (c). For the sequence pair in each sub-figure, the first one is the support, and the second one is the query.

\textbf{Attention under the CLIP(RN50)}.
For the action category ``Laying something on the table on its side, not upright" in SSv2-small,
the visualizations in \cref{fig:8}, compared with CLIP-FSAR, our CLIP-$\mathrm{M^2}$DF effectively focuses on action related backgrounds, and pays less attention to the unrelated backgrounds. Also, the attention is more accurate.   
\cref{fig:9} shows the attention visualizations of CLIP-$\mathrm{M^2}$DF on Kinetics and the action category is ``contact juggling". The attention is also more accurate than CLIP-FSAR and pays less attention to unrelated backgrounds.

\textbf{Attention under the CLIP(VIT-B/16)}.
\cref{fig:10} The action class is ``Laying something on the table on its side, not upright". We can see the attention of our model is more accurate, and the sequence attentions are more coherent.
\cref{fig:11} shows the attention visualization of our CLIP-$\mathrm{M^2}$DF on Kinetics under the 5-way 5-shot setting. The action class is ``contact juggling". The attention of our model focuses on the entire moving entity, but FSAR's attention only focuses on a portion of the moving entity. Also, the sequence attentions of our model are more coherent. 

%%%%%%%%%%%%%%%%%%%%%%%%%%%%%%%%%%%%Conclusion%%%%%%%%%%%%%%%%%%%%%%%%%%%%%%%%%%%%%%%%
%%%%%%%%%%%%%%%%%%%%%%%%%%%%%%%%%%%%%%%%%%%%%%%%%%%%%%%%%%%%%%%%%%%%%%%%%%%%%%%%%%%%%%
\section{Conclusion}
In this paper, we use CLIP as the backbone. 
Firstly, we get the probability prompt embedding for the query through the matching score and uniform sampling. 
Secondly, we merge the prompt embedding with the visual embedding and temporal context using the MMFE, through which we focus on both the local temporal context and global temporal context. Both of the views contain the features of text and visuals. 
Thirdly, we use distance fusion and mutual distillation to let the views study from each other.

\clearpage
\bibliographystyle{named}
\bibliography{main}

\begin{thebibliography}{}

\bibitem[\protect\citeauthoryear{Albanie \bgroup \em et al.\egroup }{2018}]{2018Emotion}
Samuel Albanie, Arsha Nagrani, Andrea Vedaldi, and Andrew Zisserman.
\newblock Emotion recognition in speech using cross-modal transfer in the wild.
\newblock In {\em Proceedings of the 26th ACM international conference on Multimedia}, pages 292--301, 2018.

\bibitem[\protect\citeauthoryear{Cao \bgroup \em et al.\egroup }{2020}]{cao2020few-otam}
Kaidi Cao, Jingwei Ji, Zhangjie Cao, Chien-Yi Chang, and Juan~Carlos Niebles.
\newblock Few-shot video classification via temporal alignment.
\newblock In {\em Proceedings of the IEEE/CVF Conference on Computer Vision and Pattern Recognition}, pages 10618--10627, 2020.

\bibitem[\protect\citeauthoryear{Carreira and Zisserman}{2017}]{carreira2017quo-kinetics}
Joao Carreira and Andrew Zisserman.
\newblock Quo vadis, action recognition? a new model and the kinetics dataset.
\newblock In {\em proceedings of the IEEE Conference on Computer Vision and Pattern Recognition}, pages 6299--6308, 2017.

\bibitem[\protect\citeauthoryear{Chen \bgroup \em et al.\egroup }{2018}]{chen1804semantic-few-augmentation3}
Z~Chen, Y~Fu, Y~Zhang, YG~Jiang, X~Xue, and L~Sigal.
\newblock Semantic feature augmentation in few-shot learning. arxiv 2018.
\newblock {\em arXiv preprint arXiv:1804.05298}, 2018.

\bibitem[\protect\citeauthoryear{Finn \bgroup \em et al.\egroup }{2017}]{finn2017model-MAML-few-shot-model1}
Chelsea Finn, Pieter Abbeel, and Sergey Levine.
\newblock Model-agnostic meta-learning for fast adaptation of deep networks.
\newblock In {\em International conference on machine learning}, pages 1126--1135. PMLR, 2017.

\bibitem[\protect\citeauthoryear{Fu \bgroup \em et al.\egroup }{2020}]{fu2020depth}
Yuqian Fu, Li~Zhang, Junke Wang, Yanwei Fu, and Yu-Gang Jiang.
\newblock Depth guided adaptive meta-fusion network for few-shot video recognition.
\newblock In {\em Proceedings of the 28th ACM International Conference on Multimedia}, pages 1142--1151, 2020.

\bibitem[\protect\citeauthoryear{Goyal \bgroup \em et al.\egroup }{2017}]{goyal2017something}
Raghav Goyal, Samira Ebrahimi~Kahou, Vincent Michalski, Joanna Materzynska, Susanne Westphal, Heuna Kim, Valentin Haenel, Ingo Fruend, Peter Yianilos, Moritz Mueller-Freitag, et~al.
\newblock The" something something" video database for learning and evaluating visual common sense.
\newblock In {\em Proceedings of the IEEE international conference on computer vision}, pages 5842--5850, 2017.

\bibitem[\protect\citeauthoryear{Kingma and Ba}{2014}]{2014Adam}
Diederik Kingma and Jimmy Ba.
\newblock Adam: A method for stochastic optimization.
\newblock {\em Computer Science}, 2014.

\bibitem[\protect\citeauthoryear{Kuehne \bgroup \em et al.\egroup }{2011}]{kuehne2011hmdb}
Hildegard Kuehne, Hueihan Jhuang, Est{\'\i}baliz Garrote, Tomaso Poggio, and Thomas Serre.
\newblock Hmdb: a large video database for human motion recognition.
\newblock In {\em 2011 International conference on computer vision}, pages 2556--2563. IEEE, 2011.

\bibitem[\protect\citeauthoryear{Lea \bgroup \em et al.\egroup }{2016}]{lea2016temporal-tcn}
Colin Lea, Rene Vidal, Austin Reiter, and Gregory~D Hager.
\newblock Temporal convolutional networks: A unified approach to action segmentation.
\newblock In {\em Computer Vision--ECCV 2016 Workshops: Amsterdam, The Netherlands, October 8-10 and 15-16, 2016, Proceedings, Part III 14}, pages 47--54. Springer, 2016.

\bibitem[\protect\citeauthoryear{Li \bgroup \em et al.\egroup }{2022}]{li2022ta2n}
Shuyuan Li, Huabin Liu, Rui Qian, Yuxi Li, John See, Mengjuan Fei, Xiaoyuan Yu, and Weiyao Lin.
\newblock Ta2n: Two-stage action alignment network for few-shot action recognition.
\newblock In {\em Proceedings of the AAAI Conference on Artificial Intelligence}, volume~36, pages 1404--1411, 2022.

\bibitem[\protect\citeauthoryear{Liu \bgroup \em et al.\egroup }{2022}]{liu2022multidimensionalMPRE}
Shuwen Liu, Min Jiang, and Jun Kong.
\newblock Multidimensional prototype refactor enhanced network for few-shot action recognition.
\newblock {\em IEEE Transactions on Circuits and Systems for Video Technology}, 32(10):6955--6966, 2022.

\bibitem[\protect\citeauthoryear{M{\"u}ller}{2007}]{muller2007dynamic}
Meinard M{\"u}ller.
\newblock Dynamic time warping.
\newblock {\em Information retrieval for music and motion}, pages 69--84, 2007.

\bibitem[\protect\citeauthoryear{Ni \bgroup \em et al.\egroup }{2022}]{ni2022Multi-modal-CLIP}
Xinzhe Ni, Hao Wen, Yong Liu, Yatai Ji, and Yujiu Yang.
\newblock Multimodal prototype-enhanced network for few-shot action recognition.
\newblock {\em arXiv preprint arXiv:2212.04873}, 2022.

\bibitem[\protect\citeauthoryear{Perez and Wang}{2017}]{perez2017effectiveness-few-augmentation1}
Luis Perez and Jason Wang.
\newblock The effectiveness of data augmentation in image classification using deep learning.
\newblock {\em arXiv preprint arXiv:1712.04621}, 2017.

\bibitem[\protect\citeauthoryear{Perrett \bgroup \em et al.\egroup }{2021}]{perrett2021temporal-trx}
Toby Perrett, Alessandro Masullo, Tilo Burghardt, Majid Mirmehdi, and Dima Damen.
\newblock Temporal-relational crosstransformers for few-shot action recognition.
\newblock In {\em Proceedings of the IEEE/CVF Conference on Computer Vision and Pattern Recognition}, pages 475--484, 2021.

\bibitem[\protect\citeauthoryear{Radford \bgroup \em et al.\egroup }{2021}]{Clip-origin2021}
Alec Radford, Jong~Wook Kim, Chris Hallacy, Aditya Ramesh, Gabriel Goh, Sandhini Agarwal, Girish Sastry, Amanda Askell, Pamela Mishkin, Jack Clark, et~al.
\newblock Learning transferable visual models from natural language supervision.
\newblock In {\em International conference on machine learning}, pages 8748--8763. PMLR, 2021.

\bibitem[\protect\citeauthoryear{Ratner \bgroup \em et al.\egroup }{2017}]{ratner2017learning-few-augmentation2}
Alexander~J Ratner, Henry Ehrenberg, Zeshan Hussain, Jared Dunnmon, and Christopher R{\'e}.
\newblock Learning to compose domain-specific transformations for data augmentation.
\newblock {\em Advances in neural information processing systems}, 30, 2017.

\bibitem[\protect\citeauthoryear{Shen \bgroup \em et al.\egroup }{2019}]{2019Amalgamating}
Chengchao Shen, Xinchao Wang, Jie Song, Li~Sun, and Mingli Song.
\newblock Amalgamating knowledge towards comprehensive classification.
\newblock {\em Proceedings of the AAAI Conference on Artificial Intelligence}, 33:3068--3075, 2019.

\bibitem[\protect\citeauthoryear{Simon \bgroup \em et al.\egroup }{2020}]{simon2020adaptive-few_shot_learning1}
Christian Simon, Piotr Koniusz, Richard Nock, and Mehrtash Harandi.
\newblock Adaptive subspaces for few-shot learning.
\newblock In {\em Proceedings of the IEEE/CVF conference on computer vision and pattern recognition}, pages 4136--4145, 2020.

\bibitem[\protect\citeauthoryear{Snell \bgroup \em et al.\egroup }{2017}]{snell2017prototypical_few_shot_learning2}
Jake Snell, Kevin Swersky, and Richard Zemel.
\newblock Prototypical networks for few-shot learning.
\newblock {\em Advances in neural information processing systems}, 30, 2017.

\bibitem[\protect\citeauthoryear{Soomro \bgroup \em et al.\egroup }{2012}]{soomro2012ucf101}
Khurram Soomro, Amir~Roshan Zamir, and Mubarak Shah.
\newblock Ucf101: A dataset of 101 human actions classes from videos in the wild.
\newblock {\em arXiv preprint arXiv:1212.0402}, 2012.

\bibitem[\protect\citeauthoryear{Sung \bgroup \em et al.\egroup }{2018}]{sung2018learning-relation-networkfew_shot_learning3}
Flood Sung, Yongxin Yang, Li~Zhang, Tao Xiang, Philip~HS Torr, and Timothy~M Hospedales.
\newblock Learning to compare: Relation network for few-shot learning.
\newblock In {\em Proceedings of the IEEE conference on computer vision and pattern recognition}, pages 1199--1208, 2018.

\bibitem[\protect\citeauthoryear{Thatipelli \bgroup \em et al.\egroup }{2022}]{thatipelli2022spatio-strm}
Anirudh Thatipelli, Sanath Narayan, Salman Khan, Rao~Muhammad Anwer, Fahad~Shahbaz Khan, and Bernard Ghanem.
\newblock Spatio-temporal relation modeling for few-shot action recognition.
\newblock In {\em Proceedings of the IEEE/CVF Conference on Computer Vision and Pattern Recognition}, pages 19958--19967, 2022.

\bibitem[\protect\citeauthoryear{Vinyals \bgroup \em et al.\egroup }{2016}]{vinyals2016matching—few_shot_learning5}
Oriol Vinyals, Charles Blundell, Timothy Lillicrap, Daan Wierstra, et~al.
\newblock Matching networks for one shot learning.
\newblock {\em Advances in neural information processing systems}, 29, 2016.

\bibitem[\protect\citeauthoryear{Wang \bgroup \em et al.\egroup }{2016}]{wang2016temporal-tsn}
Limin Wang, Yuanjun Xiong, Zhe Wang, Yu~Qiao, Dahua Lin, Xiaoou Tang, and Luc Van~Gool.
\newblock Temporal segment networks: Towards good practices for deep action recognition.
\newblock In {\em European conference on computer vision}, pages 20--36. Springer, 2016.

\bibitem[\protect\citeauthoryear{Wang \bgroup \em et al.\egroup }{2021}]{wang2021semantic}
Xiao Wang, Weirong Ye, Zhongang Qi, Xun Zhao, Guangge Wang, Ying Shan, and Hanzi Wang.
\newblock Semantic-guided relation propagation network for few-shot action recognition.
\newblock In {\em Proceedings of the 29th ACM International Conference on Multimedia}, pages 816--825, 2021.

\bibitem[\protect\citeauthoryear{Wang \bgroup \em et al.\egroup }{2022a}]{wang2022task}
Jiayi Wang, Yi~Jin, Songhe Feng, and Yidong Li.
\newblock Task adaptive modeling for few-shot action recognition.
\newblock In {\em 2022 IEEE 24th International Workshop on Multimedia Signal Processing (MMSP)}, pages 1--6. IEEE, 2022.

\bibitem[\protect\citeauthoryear{Wang \bgroup \em et al.\egroup }{2022b}]{wang2022-hybrid}
Xiang Wang, Shiwei Zhang, Zhiwu Qing, Mingqian Tang, Zhengrong Zuo, Changxin Gao, Rong Jin, and Nong Sang.
\newblock Hybrid relation guided set matching for few-shot action recognition.
\newblock In {\em Proceedings of the IEEE/CVF Conference on Computer Vision and Pattern Recognition}, pages 19948--19957, 2022.

\bibitem[\protect\citeauthoryear{Wang \bgroup \em et al.\egroup }{2023a}]{wang2023clip}
Xiang Wang, Shiwei Zhang, Jun Cen, Changxin Gao, Yingya Zhang, Deli Zhao, and Nong Sang.
\newblock Clip-guided prototype modulating for few-shot action recognition.
\newblock {\em arXiv preprint arXiv:2303.02982}, 2023.

\bibitem[\protect\citeauthoryear{Wang \bgroup \em et al.\egroup }{2023b}]{wang2023molo}
Xiang Wang, Shiwei Zhang, Zhiwu Qing, Changxin Gao, Yingya Zhang, Deli Zhao, and Nong Sang.
\newblock Molo: Motion-augmented long-short contrastive learning for few-shot action recognition.
\newblock {\em arXiv preprint arXiv:2304.00946}, 2023.

\bibitem[\protect\citeauthoryear{Wang \bgroup \em et al.\egroup }{2023c}]{2023Task-AwareDual-Representation}
Xiao Wang, Weirong Ye, Zhongang Qi, Guangge Wang, Jianping Wu, Ying Shan, Xiaohu Qie, and Hanzi Wang.
\newblock Task-aware dual-representation network for few-shot action recognition.
\newblock {\em IEEE Transactions on Circuits and Systems for Video Technology}, pages 1--1, 2023.

\bibitem[\protect\citeauthoryear{Wanyan \bgroup \em et al.\egroup }{2023}]{wanyan2023active}
Yuyang Wanyan, Xiaoshan Yang, Chaofan Chen, and Changsheng Xu.
\newblock Active exploration of multimodal complementarity for few-shot action recognition.
\newblock In {\em Proceedings of the IEEE/CVF Conference on Computer Vision and Pattern Recognition}, pages 6492--6502, 2023.

\bibitem[\protect\citeauthoryear{Wu \bgroup \em et al.\egroup }{2022}]{wu2022motion}
Jiamin Wu, Tianzhu Zhang, Zhe Zhang, Feng Wu, and Yongdong Zhang.
\newblock Motion-modulated temporal fragment alignment network for few-shot action recognition.
\newblock In {\em Proceedings of the IEEE/CVF Conference on Computer Vision and Pattern Recognition}, pages 9151--9160, 2022.

\bibitem[\protect\citeauthoryear{Yu \bgroup \em et al.\egroup }{}]{yu4104257ftan}
Bin Yu, Yonghong Hou, Zihui Guo, Zhiyi Gao, and Yueyang Li.
\newblock Ftan: Frame-to-frame temporal alignment network with contrastive learning for few-shot action recognition.
\newblock {\em Available at SSRN 4104257}.

\bibitem[\protect\citeauthoryear{Zhang \bgroup \em et al.\egroup }{2018}]{2018Deep}
Ying Zhang, Tao Xiang, Timothy~M. Hospedales, and Huchuan Lu.
\newblock Deep mutual learning.
\newblock In {\em 2018 IEEE/CVF Conference on Computer Vision and Pattern Recognition (CVPR)}, 2018.

\bibitem[\protect\citeauthoryear{Zhang \bgroup \em et al.\egroup }{2020}]{zhang2020few-ARN}
Hongguang Zhang, Li~Zhang, Xiaojuan Qi, Hongdong Li, Philip~HS Torr, and Piotr Koniusz.
\newblock Few-shot action recognition with permutation-invariant attention.
\newblock In {\em Computer Vision--ECCV 2020: 16th European Conference, Glasgow, UK, August 23--28, 2020, Proceedings, Part V 16}, pages 525--542. Springer, 2020.

\bibitem[\protect\citeauthoryear{Zheng \bgroup \em et al.\egroup }{2022}]{zheng2022few-hcl}
Sipeng Zheng, Shizhe Chen, and Qin Jin.
\newblock Few-shot action recognition with hierarchical matching and contrastive learning.
\newblock In {\em European Conference on Computer Vision}, pages 297--313. Springer, 2022.

\bibitem[\protect\citeauthoryear{Zhu and Yang}{2018}]{zhu2018compound}
Linchao Zhu and Yi~Yang.
\newblock Compound memory networks for few-shot video classification.
\newblock In {\em Proceedings of the European Conference on Computer Vision (ECCV)}, pages 751--766, 2018.

\bibitem[\protect\citeauthoryear{Zhu and Yang}{2020}]{zhu2020label}
Linchao Zhu and Yi~Yang.
\newblock Label independent memory for semi-supervised few-shot video classification.
\newblock {\em IEEE Transactions on Pattern Analysis and Machine Intelligence}, 44(1):273--285, 2020.

\bibitem[\protect\citeauthoryear{Zhu \bgroup \em et al.\egroup }{2021}]{zhu2021few-PAL}
Xiatian Zhu, Antoine Toisoul, Juan-Manuel Perez-Rua, Li~Zhang, Brais Martinez, and Tao Xiang.
\newblock Few-shot action recognition with prototype-centered attentive learning.
\newblock {\em arXiv preprint arXiv:2101.08085}, 2021.

\end{thebibliography}
%%%%%%%%%%%%%%%%%%%%%%%%%%%%%%%%%%%%%%%%%%%%%%%%%%%%%%%%%%%%%%%%%%%%%%%%%%%%%%

\end{document}